\documentclass[letterpaper]{article} 
\usepackage[preprint]{aaai2027}
\usepackage[hyphens]{url}  
\usepackage{graphicx} 
\usepackage{natbib}  
\usepackage{caption} 
\usepackage{booktabs}

\usepackage{amsmath}
\usepackage{amssymb}
\usepackage{mathtools}
\usepackage{amsthm}

\theoremstyle{plain}
\newtheorem{theorem}{Theorem}
\newtheorem{proposition}{Proposition}
\newtheorem{lemma}{Lemma}
\newtheorem{corollary}{Corollary}
\theoremstyle{definition}

\newtheorem{assumption}{Assumption}
\theoremstyle{remark}
\newtheorem{remark}{Remark}

\title{On the Rate of Convergence of Kolmogorov--Arnold Network Regression Estimators}
\author{
    Wei Liu\textsuperscript{\rm 1,\rm 3}\corresponding,
    Eleni Chatzi\textsuperscript{\rm 2},
    Zhilu Lai\textsuperscript{\rm 3}
}
\affiliations{
    \textsuperscript{\rm 1}CNRS@CREATE, Singapore\\
    \textsuperscript{\rm 2}Department of Civil, Environmental and Geomatic Engineering, ETH Z\"urich\\
    \textsuperscript{\rm 3}Internet of Things Thrust,
    The Hong Kong University of Science and Technology (Guangzhou)\\
    liu.wei@cnrsatcreate.sg, chatzi@ibk.baug.ethz.ch, zhilulai@hkust-gz.edu.cn
}

\begin{document}

\maketitle

\begin{abstract}
Kolmogorov--Arnold Networks (KANs) approximate multivariate functions by composing univariate
transformations through additive or multiplicative aggregation. We establish convergence guarantees
for KANs whose univariate components are B-splines. The least-squares estimator over the KAN spline
sieve attains the rate \(O((\log n / n)^{2r/(2r+1)})\), uniformly over a ball of regression functions
admitting a KAN representation with univariate components of Sobolev smoothness \(r\); a matching
lower bound of order \(n^{-2r/(2r+1)}\) shows this is minimax optimal up to the logarithmic factor,
which we trace to the nonlinearity of the sieve rather than to the architecture. The rate is free of
the ambient dimension \(d\); this dimension-free exponent reflects the assumed KAN structure of the
target, not an escape from the minimax rate \(n^{-2r/(2r+d)}\) on Sobolev classes over \([0,1]^d\). We derive a
knot-selection rule, show that penalized selection over a dyadic knot grid attains the rate adaptively
in the unknown smoothness, and show that univariate components are not identifiable under
centering alone, so consistency of the fit does not imply consistency of the components. On targets of
exactly known smoothness the fitted risk exponent is at least as steep as the bound in every
configuration, and the predicted knot scaling and $k^{-r}$ approximation decay are checked directly.
\end{abstract}


\section{Introduction}

Convergence rates for nonparametric regression with neural networks have been studied extensively
\citep{barron2002universal,barron1994approximation,imaizumi2019deep,10.1214/18-AOS1747,10.1214/19-AOS1875}.
\citet{stone1982optimal} established the minimax-optimal rate for general nonparametric regression,
which remains the benchmark against which modern architectures are judged, and deep fully connected
networks are now known to approach it: \citet{kohler2021rate} obtain
\(O((\log n)^6 n^{-2r/(2r+K)})\) under suitable smoothness. Neural networks are thus competitive with
minimax-optimal procedures, but their entangled representations lack explicit structure, which limits
interpretability and complicates analysis. Spline-based methods reach the same optimality by a
structured and well-understood route: B-spline estimators attain \(O(n^{-2r/(2r+1)})\)
\citep{speckman1985spline,nussbaum1985spline,he1994convergence,li1995global}, and more generally sieve
estimators are minimax optimal when the basis is matched to the class \citep{chen2007large}, as are
kernel and local polynomial methods under suitable bandwidths \citep{gyorfi2002distribution}. What
these share is that model structure is linked directly to approximation properties.

Kolmogorov--Arnold Networks \citep{liu2024kan} sit at the intersection of these two lines: they are
layered compositions, like neural networks, but built from univariate basis expansions, like splines.
Their foundation is the Kolmogorov--Arnold representation theorem
\citep{kolmogorov1957representations,arnol1957functions,schmidt2021kolmogorov}, which writes any
multivariate continuous function as a finite sum of univariate functions of sums of univariate
functions; later variants admit multiplicative and hybrid aggregation \citep{liu2024kan2}. Despite
rapid empirical uptake
\citep{vaca2024kolmogorov,bodner2024convolutional,kiamari2024gkan,koenig2024kan,xu2024kolmogorov},
formal convergence analyses remain scarce. \citet{gao2025convergence} study optimization and
generalization of KANs trained by SGD; our question is complementary and purely statistical --- what
rate does the empirical risk minimizer over a KAN spline sieve achieve, and is it optimal?

\paragraph{Contributions.}
\begin{itemize}
\item We analyze the least-squares estimator over the KAN spline sieve and prove a rate of
$O((\log n/n)^{2r/(2r+1)})$, uniformly over a ball of KAN functions with univariate components of
Sobolev smoothness $r$ (Theorems~\ref{thm:kan_add} and~\ref{thm:kan_mult}). The analysis makes
explicit three points that are easy to get wrong: the outer functions must be controlled in the
\emph{sup-norm} Sobolev scale, because their approximation error is incurred at the random point
$T_q(X)$ (Remark~\ref{rem:outer_sup}); their smoothness budget must be measured on a
\emph{scale-free} footing, since their domain is not of unit length and an unnormalized bound
silently inflates the risk (Remark~\ref{rem:scalefree});
and the sieve must carry explicit boundedness constraints, without which the estimator is not even
consistent in practice (Remark~\ref{rem:bounded}).
\item We give a matching minimax lower bound of order $n^{-2r/(2r+1)}$ over the same ball
(Corollary~\ref{cor:minimax}), so the estimator is optimal up to the logarithmic factor. We trace
that factor to the \emph{nonlinearity} of the sieve, show it disappears for linear sub-models, and
give a local-entropy condition under which it disappears in general (Remark~\ref{rem:log},
Proposition~\ref{prop:nolog}).
\item We show that centering does \emph{not} identify a KAN representation --- a one-parameter scale
group survives, along with further ambiguities when $d = 1$ or $Q \ge 2$
(Proposition~\ref{prop:noident}) --- and prove a positive identifiability result for a single additive
unit with $d \ge 2$ via Pexider's functional equation (Proposition~\ref{prop:ident}). Consequently
convergence of $\hat f_n$ does \emph{not} imply convergence of the fitted components
(Remark~\ref{rem:components}).
\item We derive the knot rule $k_n \asymp (n/\log n)^{1/(2r+1)}$ (Corollary~\ref{cor:knot}) and prove
that penalized selection over a dyadic knot grid attains the rate \emph{adaptively}, without knowing
$r$, for every $r$ up to the spline order (Theorem~\ref{thm:adaptive},
Remark~\ref{rem:adaptive_cap}).
\item We verify the theory on targets of exactly known smoothness (Section~\ref{sec:sim}): the risk
decays no slower than predicted in every configuration, with the ordering in $r$ the theory requires;
the univariate approximation error decays at the predicted $k^{-r}$; and the optimal knot number, the
adaptivity of cross-validation, robustness across a tenfold noise range and --- most tellingly --- the
dimension-free exponent are all borne out, the last while two structure-free baselines on the same
data degrade toward the curse-of-dimensionality rate as $d$ grows.
\end{itemize}

\paragraph{Relation to prior rate results.}
The building blocks --- spline approximation of Sobolev functions, empirical-process bounds for sieve
least squares --- are classical
\citep{stone1982optimal,speckman1985spline,chen2007large,gyorfi2002distribution}. Our contribution is
not a new proof technique but the transfer of these tools to the KAN class, together with three things
specific to it that do \emph{not} reduce to the classical theory: the outer functions must be
controlled at the random points $T_q(X)$ rather than in an unweighted $L^2$ sense
(Remark~\ref{rem:outer_sup}); multiplicative aggregation needs a telescoping argument with no
counterpart in the additive theory, and its cost enters the constant explicitly
(Theorem~\ref{thm:kan_mult}); and the parameterization is non-identifiable in a way that governs
whether the learned components converge (Propositions~\ref{prop:noident}--\ref{prop:ident}).
The closest precursor is \citet{horowitz2007rate}: their $G(\sum_j f_j(x_j))$ with unknown link is a
single additive KAN node, estimated at the univariate rate; Theorems~\ref{thm:kan_add}
and~\ref{thm:kan_mult} extend this to $Q$ nodes and multiplicative aggregation. It is
worth distinguishing our setting from the deep-network rates over anisotropic Besov spaces of
\citet{suzuki2021deep} and the ReLU analysis of \citet{10.1214/19-AOS1875}. Those concern
generic deep compositions of \emph{multivariate} functions; the KAN class --- finite sums of univariate
Sobolev components combined by addition or multiplication --- is neither contained in nor contains that
class, so neither our rates nor the knot rule follow from theirs. Adaptivity to unknown smoothness, in
the spirit of \citet{shi2024adaptive}, is addressed directly by Theorem~\ref{thm:adaptive} and
empirically in Section~\ref{sec:sim}.

\section{Setup}
\label{sec:setup}

We observe $\{(X_i,Y_i)\}_{i=1}^n$ with
\begin{equation}
Y_i = f_0(X_i) + \varepsilon_i, \qquad X_i \in [0,1]^d ,
\label{eq:model}
\end{equation}
and estimate $f_0 : [0,1]^d \to \mathbb{R}$. The domain $[0,1]^d$ is standard and costs no generality,
any bounded domain being affinely equivalent to it \citep{tsybakov2008nonparametric}. We write
$\|g\|^2 = \int g^2\,dP_X$; under Assumption~\ref{ass:design} this is equivalent to the Lebesgue
$L^2([0,1]^d)$ norm, so the rates below may be read in either. Let $W^{r,p}(\mathcal{J})$ be the
Sobolev space of functions on an interval $\mathcal{J}$ whose $r$-th weak derivative is in
$L^p(\mathcal{J})$, and $W^r = W^{r,2}$. Let $S_{k,m}(\mathcal{J})$ be the space of splines of order
$m$ (degree $m-1$) with $k$ uniform interior knots on $\mathcal{J}$, so
$\dim S_{k,m}(\mathcal{J}) = k+m$; Appendix~\ref{app:bspline} recalls the standard facts about the
B-spline basis that we use.

A KAN models $f_0$ as a finite sum of univariate nonlinearities composed with structured aggregations,
\begin{equation}
\begin{gathered}
f(x) = \sum_{q=1}^Q g_q\big( T_q(x) \big),\\
T_q(x) =
\begin{cases}
\sum_{j=1}^d \psi_{qj}(x_j), & \text{(additive)} \\[2pt]
\prod_{j=1}^d \psi_{qj}(x_j), & \text{(multiplicative)}
\end{cases}
\end{gathered}
\label{eq:kan}
\end{equation}
where $Q$ is the number of units, $g_q : \mathbb{R} \to \mathbb{R}$ are the \emph{outer} functions and
$\psi_{qj} : [0,1] \to \mathbb{R}$ the \emph{inner} ones, all univariate. Taking every node additive
gives the \emph{additive} class; allowing each node either type gives the
\emph{hybrid} class. Throughout, $r \ge 1$ is an \emph{integer} denoting the Sobolev
smoothness of these univariate components, so the embedding $W^r([0,1]) \subset C([0,1])$ makes every
composition well defined. Integrality is what lets us write the $r$-th derivative $\psi^{(r)}$ and
invoke the spline estimates of Lemma~\ref{lem:approx}; fractional smoothness would require the same
argument on a Besov scale, which we do not pursue.

\subsection{Assumptions}
\label{sec:assumptions}

\begin{assumption}[Design]
\label{ass:design}
The design distribution $P_X$ on $[0,1]^d$ has a Lebesgue density $p_X$ satisfying
$0 < c_X \le p_X(x) \le C_X < \infty$.
\end{assumption}

\begin{assumption}[Sampling and noise]
\label{ass:noise}
The pairs $\{(X_i,\varepsilon_i)\}_{i=1}^{n}$ are i.i.d., $\mathbb{E}[\varepsilon_i \mid X_i] = 0$,
and the errors are conditionally sub-Gaussian:
$\mathbb{E}[e^{\lambda \varepsilon_i} \mid X_i] \le e^{\lambda^2 \sigma^2/2}$ for all $\lambda$
(Remark~\ref{rem:indep}).
\end{assumption}

\begin{assumption}[Regularity of the regression function]
\label{ass:target}
Fix integers $r \ge 1$ and $Q, d \ge 1$, and constants $M, M_g, L, B < \infty$. Let $c_0 \ge 1$ be the
spline stability constant of Lemma~\ref{lem:approx} (depending only on the spline order $m$), put
$M' = c_0 M$, $\bar M = \max\{dM',\, (M')^{d}\}$, and let
$\mathcal{I} = [-\bar M - 1,\ \bar M + 1]$, an interval containing the range of every node input and
of every spline perturbation of it. The regression function has the form \eqref{eq:kan}, where
\begin{enumerate}
\item[(i)] $\|\psi_{qj}\|_\infty \le M$ and $\psi_{qj} \in W^{r,2}([0,1])$ with
      $\|\psi_{qj}^{(r)}\|_{L^2} \le B$;
\item[(ii)] $g_q \in W^{r,\infty}(\mathcal{I})$ with $\|g_q\|_{L^\infty(\mathcal{I})} \le M_g$,
      $\|g_q'\|_{L^\infty(\mathcal{I})} \le L$ and the \emph{scale-free} top-derivative bound
      \begin{equation}
      |\mathcal{I}|^{\,r}\, \big\|g_q^{(r)}\big\|_{L^\infty(\mathcal{I})} \ \le\ B ,
      \qquad |\mathcal{I}| = 2(\bar M + 1) .
      \label{eq:scalefree}
      \end{equation}
\end{enumerate}
We write $f_0 \in \mathcal{F}^{\mathrm{KAN}}_r(M,M_g,L,B)$ for the class of all such functions, and
$\mathcal{F}^{\mathrm{KAN},\mathrm{add}}_r$ for the subclass with every node additive. Note
$\|f_0\|_\infty \le Q M_g$.
\end{assumption}

Two features of Assumption~\ref{ass:target} are not cosmetic. Condition (ii) is stronger than
membership in $W^{r,2}$ because the composition evaluates $g_q$ at the random point $T_q(X)$, whose
law need not have a bounded density --- for a multiplicative node with $\psi_{qj} = \mathrm{id}$ and
$X$ uniform, $\prod_j X_j$ has density $(-\log u)^{d-1}/(d-1)!$, unbounded at the origin
(Remark~\ref{rem:outer_sup}). And \eqref{eq:scalefree} measures the smoothness budget of $g_q$ on a
fixed domain: $\mathcal{I}$ is not of unit length, so an unnormalized bound would leave a factor
$|\mathcal{I}|^{2r}$ in the risk (Remark~\ref{rem:scalefree}). Keeping $M_g$ separate from $M$ matters for the
same reason: identifying them would exclude even $g_q = \mathrm{id}$, needed in
Corollary~\ref{cor:minimax}.

\paragraph{The spline sieve.}
Let $m \ge \max\{r,2\}$ ($m \ge r$ for Lemma~\ref{lem:approx}; $m \ge 2$ so that splines in
$S_{k_n,m}$ are continuous and the derivative constraint below is a true Lipschitz bound,
Remark~\ref{rem:order2}), and fix $M' = c_0M$, $M_g' = c_0M_g$,
$L' = c_0L$. For a node-type pattern $\tau \in \{\mathrm{add},\mathrm{mult}\}^{Q}$ write $T_q^{\tau}$
for the corresponding aggregation and set
\begin{equation}
\mathcal{F}_n^{\tau}
=
\left\{
\sum_{q=1}^{Q} g_q\circ T^{\tau}_q
:
\begin{aligned}
&g_q \in S_{k_n,m}(\mathcal{I}),\ \|g_q\|_\infty \le M_g', \\
&\psi_{qj} \in S_{k_n,m}([0,1]), \\
&\|\psi_{qj}\|_\infty \le M',\ \|g_q'\|_\infty \le L'
\end{aligned}
\right\},
\label{eq:sieve}
\end{equation}
and $\mathcal{F}_n = \bigcup_{\tau} \mathcal{F}_n^{\tau}$. The number of free spline coefficients is
\begin{equation}
p_n := Q(d+1)(k_n + m) \asymp k_n
\qquad \text{for fixed } (Q,d,m).
\label{eq:pn}
\end{equation}
The \textbf{spline-based KAN sieve estimator} is any empirical risk minimizer
\begin{equation}
\hat{f}_n \in \arg\min_{f \in \mathcal{F}_n} \frac{1}{n}\sum_{i=1}^{n} \big(Y_i - f(X_i)\big)^2 .
\label{eq:erm}
\end{equation}
Both inner and outer units are free parameters of \eqref{eq:erm}; no access to the true components is
assumed. The results below are independent of the numerical procedure used to solve \eqref{eq:erm};
Section~\ref{sec:sim} reports the attained training loss so its quality can be judged. The sup-norm
constraints in \eqref{eq:sieve} are load-bearing in theory and in practice
(Remark~\ref{rem:bounded}).

Note that $\mathcal{F}_n$ is \emph{not} a linear space: a composition of two splines is not a spline,
and $\mathcal{F}_n$ is not closed under addition. It is a nonlinear, finite-dimensionally
parameterized sieve, and that --- not its dimension --- is the source of the logarithmic factor in
Theorem~\ref{thm:kan_add} (Remark~\ref{rem:log}). Table~\ref{tab:convergence_rates} in
Appendix~\ref{app:bspline} situates the rate we obtain among the classical ones.

\section{Main Results}
\label{sec:main}

All results are uniform over $\mathcal{F}^{\mathrm{KAN}}_r(M,M_g,L,B)$. Stating them over a class with
\emph{fixed} constants is what makes the minimax comparison meaningful: the class of all KAN functions
with univariate Sobolev components carries no uniform bound and has infinite minimax risk
(Remark~\ref{rem:ball}). Proofs are in Appendix~\ref{app:proof}, which isolates three ingredients: an
approximation bound for KAN compositions (Lemma~\ref{lem:approx}), a metric-entropy bound for the
nonlinear sieve (Lemma~\ref{lem:entropy}), and an oracle inequality for least squares
(Lemma~\ref{lem:oracle}). Appendix~\ref{app:remarks} collects the remarks deferred below.

\begin{theorem}[Additive KAN]
\label{thm:kan_add}
Let Assumptions~\ref{ass:design}--\ref{ass:target} hold with every node additive, let $m \ge \max\{r,2\}$, and
let $\hat f_n$ be the sieve estimator \eqref{eq:erm} with
\begin{equation}
k_n \asymp \left(\frac{n}{\log n}\right)^{\frac{1}{2r+1}} .
\label{eq:kn_log}
\end{equation}
Then there is $C < \infty$ depending only on $(Q,d,m,M,M_g,L,B,\sigma,c_X,C_X)$ such that for all
$n \ge 3$,
\begin{equation}
\sup_{f_0 \in \mathcal{F}^{\mathrm{KAN},\mathrm{add}}_r}
\mathbb{E}\!\left[\| \hat{f}_n - f_0 \|^2\right]
\ \le\
C \left(\frac{\log n}{n}\right)^{\frac{2r}{2r+1}} .
\end{equation}
\end{theorem}

\begin{theorem}[Hybrid KAN]
\label{thm:kan_mult}
Under Assumptions~\ref{ass:design}--\ref{ass:target}, with each node $T_q$ allowed to be \emph{either}
additive or multiplicative, $m \ge \max\{r,2\}$, and $k_n$ as in \eqref{eq:kn_log},
\begin{equation}
\sup_{f_0 \in \mathcal{F}^{\mathrm{KAN}}_r}
\mathbb{E}\!\left[\| \hat{f}_n - f_0 \|^2\right]
\ \le\
C' \left(\frac{\log n}{n}\right)^{\frac{2r}{2r+1}},
\end{equation}
where $C' = C \cdot \max\{1, (M')^{2(d-1)}\}$ with $C$ as in Theorem~\ref{thm:kan_add} and
$M' = c_0M$ the sieve bound of \eqref{eq:sieve}.
\end{theorem}

Theorem~\ref{thm:kan_mult} is stated for genuinely hybrid targets rather than purely multiplicative
ones, because a target mixing node types lies in neither extreme class and Corollary~\ref{cor:minimax}
must cover all of $\mathcal{F}^{\mathrm{KAN}}_r$ (Remark~\ref{rem:hybrid}).

\paragraph{The exponent is free of $d$.}
The exponent $2r/(2r+1)$ is the optimal rate for estimating a \emph{univariate} function in
$W^r([0,1])$ and does not involve $d$. One qualification matters: this is a statement about
$\mathcal{F}^{\mathrm{KAN}}_r$, \emph{not} about $W^r([0,1]^d)$, whose minimax rate is
$n^{-2r/(2r+d)}$ \citep{stone1982optimal} and which no estimator can beat. What removes the dimension
is the structural assumption that $f_0$ admits a KAN representation with a \emph{fixed} number $Q$ of
units, not the architecture by itself.

\paragraph{On the logarithmic factor.}
Theorem~\ref{thm:kan_add} misses the minimax rate of Corollary~\ref{cor:minimax} by
$(\log n)^{2r/(2r+1)}$. The factor enters through Lemma~\ref{lem:oracle}, which bounds the stochastic
error by $p_n\log n/n$ rather than $p_n/n$: that is what a \emph{uniform} entropy bound gives for a
nonlinear class, and it is the same phenomenon behind the logarithms in deep-network rates
\citep{10.1214/19-AOS1875,kohler2021rate}. It is an artifact of the \emph{nonlinearity} of the sieve,
not of the KAN architecture: least squares over a $p$-dimensional \emph{linear} space is an orthogonal
projection with stochastic error exactly $\sigma^2 p/n$ and no covering argument, whereas the KAN
sieve must be controlled by chaining over a covering, which at resolution $\epsilon \asymp 1/n$ costs
$p_n \log n$. Consistently, fixing the outer functions at $g_q = \mathrm{id}$ makes $\mathcal{F}_n$ a
linear space --- a classical additive model --- and the clean rate returns. Remark~\ref{rem:log}
expands on this.

\begin{proposition}[Removing the logarithmic factor]
\label{prop:nolog}
Assume in addition to Theorem~\ref{thm:kan_add} that the sieve obeys the \emph{local} entropy bound:
writing $\mathcal{F}_n(\delta) = \{ f \in \mathcal{F}_n : \|f - f_0\| \le \delta \}$, there is
$C_\star < \infty$ with
\begin{equation}
\log N\!\left(\epsilon,\ \mathcal{F}_n(\delta),\ \|\cdot\|_\infty\right)
\le C_\star\, p_n \log(\delta/\epsilon)
\label{eq:local_entropy}
\end{equation}
for $0 < \epsilon < \delta \le 2QM_g'$. Then with $k_n \asymp n^{1/(2r+1)}$ one has
$\mathbb{E}[\|\hat f_n - f_0\|^2] = O(n^{-2r/(2r+1)})$, the exact minimax rate.
\end{proposition}

The localization in \eqref{eq:local_entropy} is in $\|\cdot\|$, not $\|\cdot\|_\infty$: it is the
$L^2$ ball around $f_0$ that the argument must cover, and it is the larger of the two. The condition
holds whenever the parameterization is bi-Lipschitz from the parameter box into
$(\mathcal{F}_n, \|\cdot\|)$; by Proposition~\ref{prop:noident} it is \emph{never} injective, so the
lower half of that would have to be verified on the quotient by the invariance group
(Remark~\ref{rem:log}). We regard this as the main technical gap left open here.

\subsection{Identifiability}

Identifiability is not needed for Theorem~\ref{thm:kan_add}, since $\|\hat f_n - f_0\|$ is invariant
under reparameterization. It is exactly what is needed if the learned components are to be
\emph{interpreted}, and it governs whether they converge at all.

\begin{proposition}[Centering does not identify a KAN]
\label{prop:noident}
Let $f(x) = \sum_{q=1}^{Q} g_q(\sum_{j=1}^d \psi_{qj}(x_j))$ on $[0,1]^d$ and impose
\begin{equation}
\int_0^1 \psi_{qj}(t)\,dt = 0 \ \ (\forall j),
\qquad
\int g_q(u) \, d\mu_q(u) = 0 ,
\label{eq:centering}
\end{equation}
where $\mu_q$ is the law of $\sum_j \psi_{qj}(X_j)$. Then:
\textup{(a)} \emph{(scale invariance)} for any $a_1,\dots,a_Q \ne 0$, the substitution
$\psi_{qj} \mapsto a_q \psi_{qj}$, $g_q \mapsto g_q(\,\cdot\,/a_q)$ leaves $f$ unchanged \emph{and}
preserves both constraints, so centering alone never identifies the representation, even up to
permutation of the $Q$ units;
\textup{(b)} \emph{(reparameterization when $d=1$)} if $d=1$, then for every strictly monotone
continuous $h$ the substitution $\psi \mapsto h \circ \psi - \int_0^1 h(\psi)$,
$g \mapsto g \circ h^{-1}(\,\cdot + \int_0^1 h(\psi)\,)$ leaves $f$ and both constraints unchanged;
\textup{(c)} \emph{(mixing when $Q \ge 2$)} with $Q=2$ and $g_1 = g_2 = \mathrm{id}$ one has
$f = \sum_j (\psi_{1j} + \psi_{2j})(x_j)$, and every splitting of each $\psi_{1j}+\psi_{2j}$ into two
centered summands yields the same $f$.
Without \eqref{eq:centering} the shift family $\psi_{qj} \mapsto \psi_{qj} + c_{qj}$,
$g_q \mapsto g_q(\cdot - \sum_j c_{qj})$ also leaves $f$ unchanged.
\end{proposition}

Part (a) corrects a claim one might expect to hold: centering removes the shift ambiguity but leaves a
one-parameter scale group, so ``identifiable up to permutation'' is false. A positive result needs a
scale normalization \emph{and} a nondegeneracy condition, and then holds for a single unit.

\begin{proposition}[Identifiability of a single additive unit]
\label{prop:ident}
Let $d \ge 2$ and suppose that for a.e.\ $x \in [0,1]^d$,
$g(\sum_{j} \psi_j(x_j)) = \tilde g(\sum_{j} \tilde\psi_j(x_j))$,
where the $\psi_j, \tilde\psi_j$ are continuous, at least two of the $\psi_j$ are non-constant, and
$g, \tilde g$ are continuous and strictly monotone on the ranges of the corresponding indices. Then
there exist $a \ne 0$ and constants $b_1,\dots,b_d$ with
\begin{equation}
\tilde\psi_j = a\,\psi_j + b_j ,
\qquad
\tilde g(u) = g\!\left(\tfrac{u - \sum_j b_j}{a}\right).
\end{equation}
If in addition both representations are centered as in \eqref{eq:centering} and scale-normalized by
$\sum_j \|\psi_j\|_{L^2}^2 = \sum_j \|\tilde\psi_j\|_{L^2}^2 = 1$, then $b_j = 0$ for all $j$ and
$a \in \{-1,+1\}$: the representation is unique up to a global sign.
\end{proposition}

The hypotheses are sharp: each failure mode in Proposition~\ref{prop:noident} violates one of them
($d = 1$ in (b), affine outer functions with $Q \ge 2$ in (c)). Extending to $Q \ge 2$ requires ruling
out affine relations among the indices $T_q$, which we leave open.

\subsection{Minimax Optimality, Knots, and Adaptivity}

\begin{corollary}[Minimax rate over the KAN class]
\label{cor:minimax}
Fix $r \ge 1$ and constants with $M_g \ge \bar M + 1$, $L \ge 1$ and $B \ge |\mathcal{I}|$. For
$X$ uniform on $[0,1]^d$ and $\varepsilon \sim \mathcal{N}(0,\sigma^2)$, the minimax risk
$R_n(\mathcal{F}_r^{\mathrm{KAN}}) = \inf_{\hat f_n} \sup_{f_0}
\mathbb{E}_{f_0}\|\hat f_n - f_0\|_{L^2([0,1]^d)}^2$ over
$\mathcal{F}_r^{\mathrm{KAN}}(M,M_g,L,B)$ satisfies, for constants $0 < c \le C < \infty$ independent
of $n$,
\begin{equation}
c\, n^{-\frac{2r}{2r+1}}
\ \le\
R_n\big(\mathcal{F}_r^{\mathrm{KAN}}\big)
\ \le\
C \left(\frac{\log n}{n}\right)^{\frac{2r}{2r+1}} .
\end{equation}
Hence the estimator of Theorems~\ref{thm:kan_add}--\ref{thm:kan_mult} is minimax rate-optimal up to
$(\log n)^{2r/(2r+1)}$, and exactly minimax optimal whenever Proposition~\ref{prop:nolog} applies.
\end{corollary}

\begin{corollary}[Optimal knot number]
\label{cor:knot}
Under the assumptions of Theorem~\ref{thm:kan_add} or \ref{thm:kan_mult},
\begin{equation}
\mathbb{E}\big[\|\hat{f}_n - f_0\|^2\big]
\ \lesssim\
\underbrace{k_n^{-2r}}_{\text{approximation}} \;+\; \underbrace{\frac{k_n \log n}{n}}_{\text{estimation}} ,
\label{eq:tradeoff_main}
\end{equation}
and the two terms are balanced by \eqref{eq:kn_log}. Ignoring the logarithmic factor, the guideline is
\begin{equation}
k_n \;\asymp\; n^{1/(2r+1)} .
\label{eq:knot}
\end{equation}
Moreover, under \eqref{eq:kn_log} the best spline approximation $\psi_{qj,k_n} \in S_{k_n,m}$ to each
univariate component satisfies
$\| \psi_{qj,k_n} - \psi_{qj} \|_{L^2([0,1])}^2 = O(k_n^{-2r})
= O((\log n/n)^{2r/(2r+1)})$,
so the sieve resolves every component to within the overall risk level.
\end{corollary}

This last statement concerns the \emph{deterministic} best approximation, not the fitted component
$\hat\psi_{qj}$. It is tempting to conclude that
$\mathbb{E}\|\hat\psi_{qj} - \psi_{qj}\|^2 = O(n^{-2r/(2r+1)})$ as well, but that is false in general:
by Proposition~\ref{prop:noident} the map from components to $f$ is not injective, so a sequence of
fits with $\|\hat f_n - f_0\| \to 0$ may have $\hat\psi_{qj}$ that do not converge at any rate ---
alternating the sign $a_q = \pm 1$ already breaks it (Remark~\ref{rem:components}).

The rule \eqref{eq:knot} is for fixed finite $r$; Remark~\ref{rem:rinfty} discusses how to read the
formal limit $r \to \infty$. It uses $r$, which is unknown in practice, and the next result removes
that dependence: one data-driven estimator attains the rate simultaneously over all smoothness classes
up to the spline order.

\begin{theorem}[Adaptation to unknown smoothness]
\label{thm:adaptive}
Let Assumptions~\ref{ass:design}--\ref{ass:target} hold, let the candidate knot counts form the
dyadic grid $\mathcal{K}_n = \{2^0,\dots,2^{\lfloor \log_2 n\rfloor}\}$, and for $k \in \mathcal{K}_n$ let
$\hat f_{n,k}$ be the sieve estimator over $\mathcal{F}_{n,k}$ (the sieve \eqref{eq:sieve} with $k$
interior knots, all sharing one fixed spline order $m \ge 2$). Select
\begin{equation}
\hat k \in \arg\min_{k \in \mathcal{K}_n}
\left\{ \frac{1}{n}\sum_{i=1}^n \big(Y_i - \hat f_{n,k}(X_i)\big)^2 + \mathrm{pen}(k) \right\}
\label{eq:penalty}
\end{equation}
with $\mathrm{pen}(k) = \kappa\,\sigma^2\,p_{n,k}\log n / n$ for a sufficiently large constant
$\kappa$, where $p_{n,k} \asymp k$. Then there is $C < \infty$, independent of $r$, such that for
\emph{every} $r$ with $1 \le r \le m$ and every $f_0 \in \mathcal{F}^{\mathrm{KAN}}_r(M,M_g,L,B)$,
\begin{equation}
\mathbb{E}\!\left[\|\hat f_{n,\hat k} - f_0\|^2\right]
\ \le\ C\left(\frac{\log n}{n}\right)^{\frac{2r}{2r+1}} .
\end{equation}
Because neither the estimator nor the penalty uses $r$, the single estimator $\hat f_{n,\hat k}$ is
rate-optimal (up to the logarithmic factor) over $\{\mathcal{F}^{\mathrm{KAN}}_r\}_{1 \le r \le m}$ at
once.
\end{theorem}

The adaptation is to the smoothness, not to the noise level: the penalty is calibrated by $\sigma^2$,
which is as unknown as $r$ in practice. This is the usual state of affairs for penalized
model selection and is normally handled by substituting a consistent estimate of $\sigma^2$, or by
dispensing with the penalty and selecting $k$ by cross-validation --- which is what our experiments
do, and which uses neither $r$ nor $\sigma$ (Section~\ref{sec:sim}). In that respect the empirical
procedure is stronger than the theorem, though it is not the procedure the theorem analyses.

The restriction $r \le m$ is not an artifact of the proof: a spline of order $m$ reproduces
polynomials of degree $m-1$, so its best approximation error saturates at $O(k^{-m})$ once $r > m$, no
matter how $k$ is chosen. Adapting beyond $m$ requires selecting over the order as well, on a
two-parameter grid $(k,m)$, at the price of one further $\log n$ in the cardinality term; we state the
single-order version because it is the one our simulations implement (Remark~\ref{rem:adaptive_cap}).
The dyadic grid keeps $|\mathcal{K}_n| \le 1+\log_2 n$, so selection costs only the same $\log n$
already present in Theorem~\ref{thm:kan_add}.

\section{Simulation Study}
\label{sec:sim}

The experiments check the quantitative predictions the theorems make, not the practical merits of KANs
against competing methods. Four predictions are testable. \textbf{(P1)} The univariate spline
approximation error decays like $k^{-r}$ (Lemma~\ref{lem:approx}). \textbf{(P2)} The $L^2$ risk decays
like $n^{-2r/(2r+1)}$ up to logarithmic factors (Theorems~\ref{thm:kan_add},~\ref{thm:kan_mult}).
\textbf{(P3)} The exponent tracks $r$. \textbf{(P4)} The risk is U-shaped in the number of knots with
minimizer growing like $n^{1/(2r+1)}$ (Corollary~\ref{cor:knot}). Code and data are provided as
supplementary material. Appendix~\ref{app:experiments} gives the full protocol and further panels.

\subsection{Design}

\paragraph{Targets with exactly known smoothness.}
Testing an exponent requires a target whose smoothness is exactly known and --- less obviously ---
whose spline approximation error is of the same order as the stochastic error at the sample sizes
used. We take
\begin{equation}
\psi_r(t) = c_r \sum_{i \ge 1} a_i \sin(2\pi i t + \phi_i),
\quad
a_i = \frac{1}{i^{\,r+1/2}\log(i+1)},
\label{eq:psi_target}
\end{equation}
with $c_r$ normalizing $\|\psi_r\|_\infty = 1$ and fixed pseudo-random phases $\phi_i$. Summing
$(1+i^2)^{s}a_i^2$ shows $\psi_r \in W^{s}([0,1])$ exactly when $s \le r$, so its smoothness is
exactly $r$ (Appendix~\ref{app:experiments}).

From $\psi_r$ we build three targets on $[0,1]^d$, each normalized to zero mean and unit variance. The
\emph{additive} target $g(d^{-1/2}\sum_{j}\psi_r(x_j))$ with $g(u)=\sin(\pi u/2)$ and $d=5$, and the
\emph{univariate} target $\psi_r(x)$ with $d=1$, instantiate Theorem~\ref{thm:kan_add}. The
\emph{multiplicative} target $g(\prod_{j=1}^{d}\psi^{+}_r(x_j))$ with $d=2$ instantiates
Theorem~\ref{thm:kan_mult}; here $\psi^+_r = 1 + \tfrac12\psi_r > 0$ has the same smoothness $r$ (a
shift and scaling leave the Sobolev seminorm unchanged). Keeping the factors few and positive holds
the product away from the degenerate concentration at zero that a longer product, or a signed
zero-mean factor, would produce.
Responses are $Y_i = f_0(X_i)+\varepsilon_i$ with $X_i \sim \mathsf{Unif}([0,1]^d)$ and
$\varepsilon_i \sim \mathcal{N}(0,\sigma^2)$, $\sigma = 0.05$. Two design pitfalls that invalidate
this kind of experiment --- a piecewise-polynomial target whose true smoothness is $3/2$ rather than
$r$, and aligned phases that make the bias negligible and the rate test vacuous --- are documented in
Remark~\ref{rem:design}.

\paragraph{Estimators.}
We fit \eqref{eq:erm} with cubic ($m=4$) B-splines and the boundedness constraints
of \eqref{eq:sieve}; Appendix~\ref{app:implementation} relates the implementation to the class as
analysed. Each target is fit by the KAN whose node structure matches it, so the sieve
contains the truth without redundant nodes. Because the model is linear in the outer coefficients
given the inner units, we solve \eqref{eq:erm} by multi-start L-BFGS \citep{liu1989limited} on the
inner units interleaved with exact least-squares solves for the outer coefficients. Multiplicative
nodes are parameterized in the log domain, $\prod_j \psi_{qj} = \exp(\sum_j \log\psi_{qj})$, which
turns a badly conditioned product into a sum; this is not merely a change of variables, since it
constrains $\psi_{qj} > 0$ and so fits the positive-factor subclass of \eqref{eq:sieve} --- harmless
here because the multiplicative target is built from $\psi^+_r > 0$ and so still lies in the fitted
class. The median attained training MSE sits at the noise level $\sigma^2 = 2.5\times10^{-3}$, so what is
measured is the ERM's statistical behaviour and not an optimization artifact. Knot counts follow
\eqref{eq:kn_log} as $k_n = \mathrm{round}(c\,(n/\log n)^{1/(2r+1)})$ with $c=3$, keeping
$k_n \ge 7$ at the smallest $n$, away from the small-$k$ regime where uniform knots align poorly.
Baselines are a fully connected ReLU MLP (width 128, depth 3, mini-batch Adam, $8000$ steps), a
sine-activation network (SIREN, \citealp{sitzmann2020implicit}), $k$-nearest-neighbour regression, and
kernel ridge regression with an RBF kernel.

Every configuration is replicated over independent draws, $10$ times for the panels of
Table~\ref{tab:slopes} and $3$--$8$ times elsewhere (Appendix~\ref{app:experiments}). Risk is measured against the
\emph{noise-free} target on a fixed test set of $N = 20{,}000$ points, so it estimates
$\|\hat f_n - f_0\|^2_{L^2}$, exactly the quantity the theorems bound. Slopes are fit by ordinary
least squares to the \emph{median} test error at each $n$, with a $95\%$ bootstrap confidence interval
over replications; error bands show the inter-quartile range.

\subsection{Results}
\label{sec:results}

\begin{figure*}[!t]
    \centering
    \includegraphics[width=0.275\textwidth]{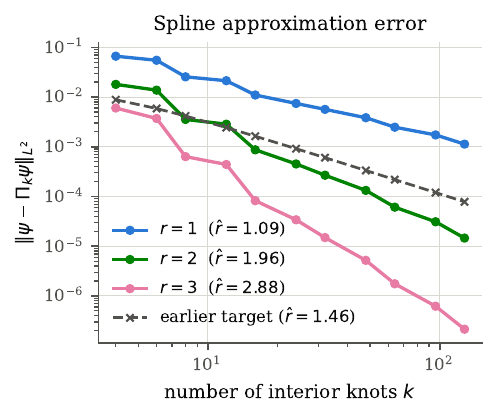}
    \hfill
    \includegraphics[width=0.29\textwidth]{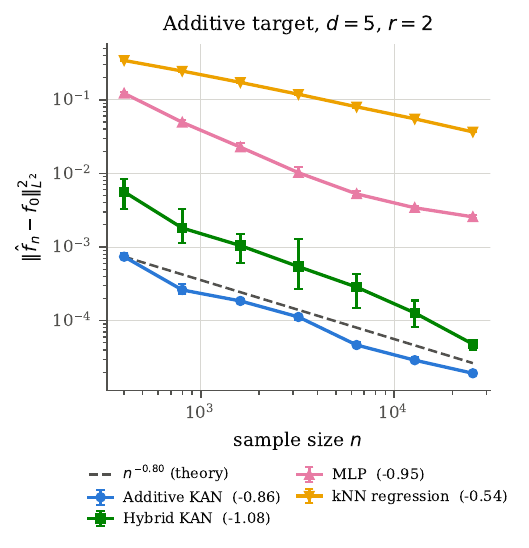}
    \hfill
    \includegraphics[width=0.29\textwidth]{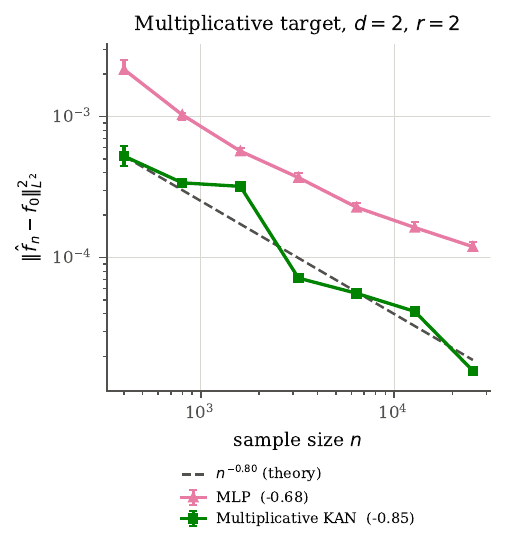}
    \caption{\textbf{Left:} measured univariate spline approximation error (P1); the fitted exponents
    $\hat r$ match the nominal smoothness $r$, while the piecewise-polynomial target of
    Remark~\ref{rem:design} (dashed) has smoothness $3/2$ rather than $2$.
    \textbf{Middle:} risk against sample size for the additive target, $d=5$, $r=2$
    (Theorem~\ref{thm:kan_add}), with the MLP and a $k$-nearest-neighbour baseline that makes no use
    of the additive structure and pays the curse of dimensionality.
    \textbf{Right:} the multiplicative target, $d=2$, $r=2$ (Theorem~\ref{thm:kan_mult}). Dashed lines
    show the reference slope $-2r/(2r+1) = -0.8$. Points are medians over $10$ replications; bands
    show the inter-quartile range.}
    \label{fig:approx}
\end{figure*}

\paragraph{P1: approximation.}
The measured exponents in Figure~\ref{fig:approx} (left), fitted over $k \ge 16$, are
$\hat r = 1.09, 1.96, 2.88$ for $r = 1,2,3$, against the predicted $1,2,3$. This isolates the univariate
input to Lemma~\ref{lem:approx} from estimation and optimization --- the composition bound is checked
in Appendix~\ref{app:experiments} --- and the shortfall at $r=3$ is a finite-$k$ effect.

\paragraph{P2: convergence rate at $r=2$.}
The KAN matched to each target attains a monotone convergence curve with fitted slope $-0.86$
(additive, $d=5$), $-0.87$ (univariate Fourier, $d=1$) and $-0.85$ (multiplicative, $d=2$); see
Table~\ref{tab:slopes}. The comparison to $-2r/(2r+1)=-0.8$ is a comparison to an \emph{upper}
bound, and the asymmetry matters: a markedly shallower slope over a range this long would be hard
to reconcile with Theorems~\ref{thm:kan_add}--\ref{thm:kan_mult}, whereas a steeper one is permitted
by them and predicted by nothing. All three
slopes are steeper than $-0.8$ and than the log-corrected bound (effective slope $\approx -0.70$ over
$n \in [400,\,2.56\times10^4]$). Two of the three bootstrap intervals nevertheless reach above $-0.8$
--- $[-0.890,-0.526]$ (multiplicative) and $[-0.937,-0.761]$ (Fourier) --- so what these sample sizes
exclude is a \emph{markedly} shallower decay, not every slope above the bound. The multiplicative
node is the one non-convex piece; a minority of starts stall, widening its interval.

\paragraph{P3: dependence on smoothness.}
The fitted slopes for $r=1,2,3$ are $-0.81,\,-0.86,\,-1.01$ (Figure~\ref{fig:smoothness}): the rate
steepens with $r$, in the correct order, but should not be read quantitatively
(Appendix~\ref{app:experiments}); the clean quantitative test of the $r$-dependence is P1.

\begin{figure*}[!t]
    \centering
    \includegraphics[width=0.624\textwidth]{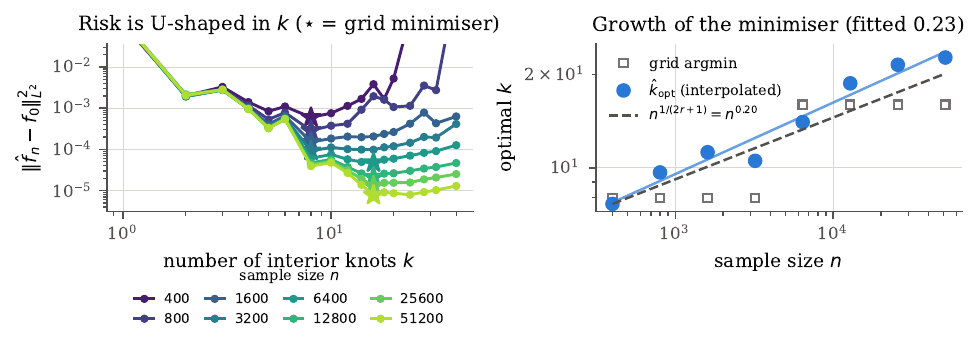}
    \hspace{0.02\textwidth}
    \includegraphics[width=0.323\textwidth]{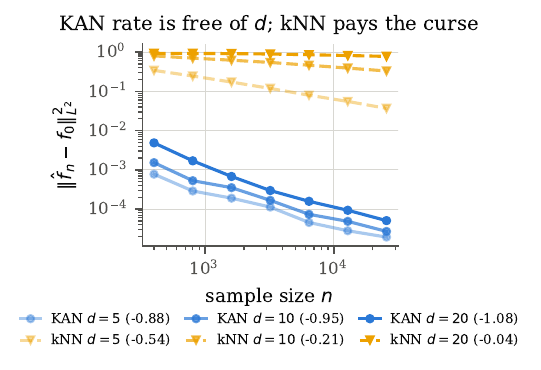}
    \caption{\textbf{Left and middle:} knot selection (P4, Corollary~\ref{cor:knot}), additive target
    with $r=2$, $n$ up to $51{,}200$ and $k$ up to $40$: an interior minimum at every $n$. Because the
    risk oscillates between grid points the grid argmin ($\star$; open squares, middle) takes only two
    values, while the minimiser from local quadratic interpolation (filled) grows at $n^{0.23}$, against
    a predicted $n^{0.2}$ that its bootstrap interval marginally excludes (see text). The left axis is clipped, so a few large-$k$ points at the smallest $n$ are
    off-scale. \textbf{Right:} the dimension-free rate. Additive target with $d \in \{5,10,20\}$: the
    KAN slopes stay at or beyond the univariate frontier ($-0.88,-0.95,-1.08$) instead of degrading,
    whereas the $k$-NN baseline degrades past the curse-of-dimensionality rate $-2r/(2r+d)$
    ($-0.54,-0.21,-0.04$ against $-0.44,-0.29,-0.17$), becoming nearly flat at $d=20$. The mild
    steepening of the KAN slopes is a finite-sample effect predicted by Corollary~\ref{cor:knot}
    (Appendix~\ref{app:experiments}), not a $d$-dependent exponent.}
    \label{fig:knots}
\end{figure*}

\paragraph{P4: knot selection.}
Figure~\ref{fig:knots} shows the predicted U-shape at every sample size: the risk falls while
approximation error dominates and rises once the estimation term takes
over. Locating the minimum needs care: the risk is not monotone \emph{between} grid points --- uniform
knots align with a target's features at some $k$ and badly at others --- so the argmin snaps to
the nearest resonant $k$. We
therefore fit a parabola to $\log$ risk against $\log k$ over the basin (the $k$ within a factor $3$
of the smallest risk). So measured, the minimiser moves from $7.6$ at $n=400$ to
$22.7$ at $n=51{,}200$, an exponent of $0.23$, $95\%$ bootstrap interval $[0.21,0.31]$; the grid argmin
gives $0.19$. The two bracket the $0.2$ of \eqref{eq:knot} (and the $0.18$ of \eqref{eq:kn_log} over
this range), the interpolated one marginally excluding both, so the agreement is ordinal, not sharp.
The schedule $k_n$ used in the convergence panels sits below the measured minimiser
at every $n$, by $1.05$--$1.54\times$, costing $1.5\times$ in risk at $n=25{,}600$ --- the harmless
direction for an upper bound. The right
arm of the U flattens as $n$ grows, so under-resolving remains the more damaging error.

\paragraph{Adapting to unknown smoothness.}
Because the risk is U-shaped in $k$ with a well-separated minimum, cross-validation can locate it
without using $r$. Selecting $k$ by $3$-fold CV yields a KAN that is never more than
$24\%$ worse than the oracle rule that knows $r$, and $31\%$ better at the largest $n$ (fitted slopes
$-0.98$ against $-0.90$, from four points); the
CV-selected $k$ grows from $5$ to $16$ over the sweep, an exponent of $0.27$ against the predicted
$0.2$, or $0.17$ with the logarithm kept (Figure~\ref{fig:adaptive}). At four sample sizes and integer
$k$ that agreement is qualitative, but the selection plainly tracks $n$. This is the empirical counterpart of
Theorem~\ref{thm:adaptive}, run at $r=2$; with $m=4$ it lies inside the range $r \le m$ where the
theorem guarantees adaptation, as do the $r \in \{1,2,3\}$ used elsewhere.

\paragraph{Structure, not locality: the curse of dimensionality.}
\looseness=-1
The dimension-free exponent is the paper's central claim, best tested against methods that pay the
curse. We add two structure-free baselines --- $k$-NN regression, a local
averaging method, and kernel ridge with an RBF kernel, a global smoother --- so the contrast cannot be
put down to locality as such. Over $W^r([0,1]^5)$ both
are minimax only at $n^{-2r/(2r+5)} = n^{-0.44}$; the $k$-NN slope is $-0.54$, with an absolute error
two to three orders of magnitude above the KAN. The sharper test is the dimension sweep
(Figure~\ref{fig:knots}, right): as $d$ grows from $5$ to $20$ the additive-KAN exponent stays steep
($-0.88,-0.95,-1.08$) and does \emph{not} drift toward the curse, while $k$-NN collapses to a nearly
flat curve, degrading at least as fast as $-2r/(2r+d)$.
Kernel ridge is the more informative of the two, because at $d=5$ it does \emph{not} look
handicapped. Refitted on the sample sizes all three share --- which shifts the slopes just quoted by
a few hundredths --- it starts at $-0.95$ against the KAN's $-0.94$, and by $d=20$ has lost most of
that exponent, $-0.18$ against the structure-free prediction $-0.17$, while the KAN's moves the other
way, to $-1.14$ (Figure~\ref{fig:dimslopes}). An estimator indistinguishable from the KAN in five
dimensions and collapsed by twenty is what separates the two explanations. The KAN rate is free of $d$ \emph{because the target has KAN
structure}, which a structure-free estimator cannot exploit --- not because the problem is
intrinsically low-dimensional.

\paragraph{Comparison with the neural baselines.}
\looseness=-1
Deep ReLU networks attain $n^{-2r/(2r+1)}$ up to logarithms on composition structures of this
kind \citep{10.1214/19-AOS1875,kohler2021rate,10.1214/18-AOS1747}, but for capacity-controlled
\emph{sequences} of networks, not one architecture at one budget, so they predict
nothing about the MLP we fit; the comparison is descriptive. The fitted slopes are
of the same order on every target --- $-0.86$ against $-0.95$ (additive), $-0.87$ against $-0.77$
(Fourier), $-0.85$ against $-0.68$ (multiplicative), with no systematic ordering --- while the risks
are separated by a median MLP/KAN factor of $123$ (additive, $d=5$), $4.1$ (multiplicative) and
$1.4$ (Fourier). That factor moves over $[91,189]$ while the risk itself
falls $38$-fold, so the separation is overwhelmingly one of constants; its
drift is exactly the $0.09$ gap between the two slopes, which over $1.8$ decades cannot be told apart
from a constant. We attribute it to the structural prior: the sieve is handed the additive form of
$f_0$ and estimates $d$ univariate functions, whereas the network must discover it.
The logarithm of Theorem~\ref{thm:kan_add} is not an alternative
explanation: it inflates the KAN's own bound by $(\log n)^{0.8} \approx 6$ at the largest $n$, an order short
of $123$, and in the wrong direction. Kernel ridge and a tuned SIREN behave the same way: no rate
deficit ($-0.91$ and $-1.20$ against $-0.86$), a median $166\times$ and $47\times$ more risk
(Appendix~\ref{app:baselines}).

Across a tenfold range of $\sigma$ the fitted slopes ($-0.88$ to $-1.01$) stay steeper
than $-0.8$, the constant growing as $\sigma^{2.0}$ (Figure~\ref{fig:noise}); the monotone drift with
$\sigma$, like that with $d$, is what \eqref{eq:tradeoff_main} predicts, not an exponent
(Appendix~\ref{app:experiments}).

\begin{table}[t]
\centering
\caption{Fitted log--log slopes of the median $\|\hat f_n-f_0\|^2_{L^2}$ against $n$, with $95\%$
bootstrap intervals over $10$ replications. The predicted exponent is $-2r/(2r+1)$ for each
KAN whose class contains the target ($-0.80$ at $r=2$) and the structure-free $-2r/(2r+d)=-0.44$ for $k$-NN on
$[0,1]^5$ --- a worst-case rate over $W^r([0,1]^5)$, not a prediction for this (far from worst-case)
target. No exponent is predicted for the MLP, whose cited rates apply to capacity-controlled
sequences of networks, not one fixed budget. Every fitted KAN slope is
steeper than $-0.8$; two of the three intervals at $r=2$ reach above it (see text). Table~\ref{tab:slopes_full} adds the predicted exponents.}
\label{tab:slopes}
\footnotesize
\setlength{\tabcolsep}{4.5pt}
\begin{tabular}{llr}
\toprule
Target & Estimator & Slope (95\% CI) \\
\midrule
additive $d=5$ & Additive KAN & $-0.86$ $[-0.92, -0.83]$ \\
additive $d=5$ & Hybrid KAN & $-1.08$ $[-1.18, -0.94]$ \\
additive $d=5$ & MLP & $-0.95$ $[-0.97, -0.92]$ \\
additive $d=5$ & kNN regression & $-0.54$ $[-0.55, -0.53]$ \\
Fourier $d=1$ & Additive KAN & $-0.87$ $[-0.94, -0.76]$ \\
Fourier $d=1$ & MLP & $-0.77$ $[-0.81, -0.73]$ \\
mult.\ $d=2$ & Multiplicative KAN & $-0.85$ $[-0.89, -0.53]$ \\
mult.\ $d=2$ & MLP & $-0.68$ $[-0.70, -0.67]$ \\
additive, $r=1$ & Additive KAN & $-0.81$ $[-0.90, -0.79]$ \\
additive, $r=2$ & Additive KAN & $-0.86$ $[-0.92, -0.83]$ \\
additive, $r=3$ & Additive KAN & $-1.01$ $[-1.04, -0.96]$ \\
\bottomrule
\end{tabular}
\end{table}

\section{Conclusion}

The least-squares estimator over the KAN spline sieve attains $O((\log n/n)^{2r/(2r+1)})$ uniformly
over a ball of KAN functions of univariate smoothness $r$, against a matching lower bound
$n^{-2r/(2r+1)}$; penalized selection over a dyadic knot grid attains it without knowing $r$. The simulations
agree with the guarantee on every target, confirm the $r$-dependent approximation decay, and locate
the optimal knot number about where predicted.

Several directions remain open. Closing the logarithmic gap between the upper and lower bounds calls
for a local-entropy analysis of the kind formulated in Proposition~\ref{prop:nolog}. Recovering the
univariate components, rather than the regression function, is a different problem, one that
identifiability conditions in the spirit of Proposition~\ref{prop:ident} make well
posed. Extending the analysis from the empirical risk minimizer to gradient-based
training \citep{gao2025convergence}, and to deeper compositions, are natural next steps.

\clearpage
\appendix
\suppressfloats[t]

\numberwithin{equation}{section}
\numberwithin{figure}{section}
\numberwithin{table}{section}
\numberwithin{theorem}{section}
\numberwithin{proposition}{section}
\numberwithin{lemma}{section}
\numberwithin{corollary}{section}
\numberwithin{definition}{section}
\numberwithin{assumption}{section}
\numberwithin{remark}{section}

\section{B-Spline Background}
\label{app:bspline}

A univariate B-spline of order $m$ (polynomial degree $m-1$) is a linear combination of basis
functions $\{B_{i,m-1}\}$ associated with a non-decreasing knot sequence. Each $B_{i,p}$ is a
piecewise polynomial with continuous derivatives up to order $p-1$ \citep{schumaker2007spline}, and
has local support, being nonzero only over $p+1$ consecutive knots, so modifying one coefficient
changes the spline only on a subinterval. The basis is generated by the Cox--de Boor recursion
\begin{align}
B_{i,0}(x) &=
\begin{cases}
1, & t_i \le x < t_{i+1}, \\
0, & \text{otherwise},
\end{cases} \\
B_{i,p}(x) &= \frac{x - t_i}{t_{i+p} - t_i} B_{i,p-1}(x) \notag \\
&\quad + \frac{t_{i+p+1} - x}{t_{i+p+1} - t_{i+1}} B_{i+1,p-1}(x)
\end{align}
for $p \ge 1$, division by zero being read as zero. Two properties are used throughout the proofs: the
basis forms a \emph{partition of unity}, $\sum_i B_{i,p}(x) = 1$, which gives
$\|\sum_i c_i B_i\|_\infty \le \max_i|c_i|$; and it is $L^\infty$-\emph{stable}, so conversely
$\max_i |c_i| \le D_m \|\sum_i c_i B_i\|_\infty$ with $D_m$ depending only on $m$
\citep[Ch.~XI]{deboor2001splines}. Both are used in Lemma~\ref{lem:entropy}.

For the classical theory, recall that over a Sobolev ball $\{f \in W^r([0,1]) : \|f\|_{W^r} \le C\}$
the minimax risk is of order $n^{-2r/(2r+1)}$ \citep{stone1982optimal}, and that it is attained by a
least-squares spline estimator $\hat f_n = \sum_i \hat c_i B_{i,p}$ with $p \ge r$ and the number of
basis functions balanced against $n$ \citep{speckman1985spline,he1994convergence,li1995global}. In
$d$ dimensions the same balance over $W^r([0,1]^d)$ gives $n^{-2r/(2r+d)}$, the curse of
dimensionality. The point of the present paper is that a KAN target, being built from univariate
components, is estimated at the \emph{univariate} exponent regardless of $d$.

\begin{table}[t]
\centering
\caption{Convergence rates for nonparametric estimators; $r$ is smoothness, $d$ the input dimension.
The rows are \emph{not} directly comparable: each is stated over a different class. The B-spline and
KAN rows concern structured classes built from univariate components of smoothness $r$, the sieve and
kernel rows concern $W^r([0,1]^d)$, and the deep-network row a composition class of effective
dimension $K$. A dimension-free exponent always reflects assumed structure, never an escape from the
lower bound of \citet{stone1982optimal} on $W^r([0,1]^d)$.}
\label{tab:convergence_rates}
\small
\begin{tabular}{@{}ll@{}}
\toprule
\textbf{Method} & \textbf{Rate} \\
\midrule
B-splines \citep{speckman1985spline} & $O(n^{-\frac{2r}{2r+1}})$ \\
Sieves \citep{chen2007large} & $O(n^{-\frac{2r}{2r+d}})$ \\
Kernel regression \citep{gyorfi2002distribution} & $O(n^{-\frac{2r}{2r+d}})$ \\
Deep nets \citep{kohler2021rate} & $O((\log n)^6 n^{-\frac{2r}{2r+K}})$ \\
KANs (Thms.~\ref{thm:kan_add},~\ref{thm:kan_mult}) & $O\big((\tfrac{\log n}{n})^{\frac{2r}{2r+1}}\big)$ \\
\bottomrule
\end{tabular}
\end{table}

\section{Deferred Remarks}
\label{app:remarks}

\begin{remark}[Why the outer functions are measured in the sup-norm scale]
\label{rem:outer_sup}
Condition (ii) of Assumption~\ref{ass:target} is stronger than membership in $W^{r,2}$, and the
strengthening is necessary rather than cosmetic. The composition $g_q \circ T_q$ evaluates $g_q$ at
the random point $T_q(X)$, so the approximation of $g_q$ must be controlled \emph{on the range of
$T_q$}. Writing $\mu_q$ for the law of $T_q(X)$,
\[
\big\| g_{q,k}(T_q) - g_q(T_q) \big\|_{L^2(P_X)}^2
= \int \big| g_{q,k} - g_q \big|^2 \, d\mu_q ,
\]
which is controlled by $\|g_{q,k} - g_q\|^2_{L^2(du)}$ only if $\mu_q$ has a bounded Lebesgue density.
That can fail: for a multiplicative node with $\psi_{qj} = \mathrm{id}$ and $X$ uniform,
$\prod_{j=1}^d X_j$ has density $(-\log u)^{d-1}/(d-1)!$ on $(0,1)$, unbounded at the origin for
$d \ge 2$. Assuming $g_q \in W^{r,\infty}$ sidesteps the issue entirely, since then
$\|g_q - g_{q,k}\|_{\infty} = O(k^{-r})$ whatever $\mu_q$ is. All results hold verbatim under the
alternative pair of assumptions ``$g_q \in W^{r,2}$ and each $\mu_q$ has a density bounded above''.
Note also that $W^{r,\infty}(\mathcal{I}) \subset W^{r,2}(\mathcal{I})$ on the bounded interval
$\mathcal{I}$, so (ii) is a genuine restriction of the class, not a change of scale.
Remark~\ref{rem:gap} records the step of the proof that makes this necessary.
\end{remark}

\begin{remark}[Why the top-derivative bound is normalized by $|\mathcal{I}|^r$]
\label{rem:scalefree}
The domain $\mathcal{I}$ of the outer functions is not of unit length. Spline approximation on an
interval $\mathcal{J}$ with $k$ uniform knots incurs an error $c\,(|\mathcal{J}|/k)^r\|g^{(r)}\|$, so
an unnormalized bound $\|g_q^{(r)}\|_\infty \le B$ would leave a factor $|\mathcal{I}|^{\,r}$ in the
approximation error and hence $|\mathcal{I}|^{2r}$ in the risk --- a dependence that is easy to
overlook.
Condition \eqref{eq:scalefree} is exactly the statement that the rescaled outer function
$\tilde g_q(u) := g_q(\tfrac{1}{2}|\mathcal{I}|\,u)$, $u \in [-1,1]$, has
$\|\tilde g_q^{(r)}\|_\infty \le 2^{-r}B$; the smoothness budget is then measured on a fixed domain
and the $|\mathcal{I}|^{\,r}$ factor cancels in Lemma~\ref{lem:approx}. If one prefers the
unnormalized bound $\|g_q^{(r)}\|_\infty \le B_0$, every result holds with $B$ replaced by
$|\mathcal{I}|^{\,r} B_0$. Only the top derivative is rescaled: the bounds $M_g$ on $\|g_q\|_\infty$
and $L$ on $\|g_q'\|_\infty$ are genuine constraints on the function itself and are used as such in
Lemma~\ref{lem:entropy} and in Step~2 of Lemma~\ref{lem:approx}.
\end{remark}

\begin{remark}[The boundedness constraints are not cosmetic]
\label{rem:bounded}
The sup-norm constraints in \eqref{eq:sieve} are what make $\mathcal{F}_n$ a uniformly bounded class,
and they are used twice: to obtain the metric-entropy bound of Lemma~\ref{lem:entropy} with a constant
free of $k_n$, and to apply the oracle inequality of Lemma~\ref{lem:oracle}. They also matter in
practice. The empirical distribution of a node input $T_q(X)$ typically leaves some cells of the outer
knot grid with little or no data; a completely unregularized least-squares solve then places
arbitrarily large coefficients on the corresponding basis functions. This can leave the training loss
unchanged, or even lower it, while inflating the $L^2$ risk by orders of magnitude.

To establish what the constraints actually do in the runs reported here, we refit every configuration
twice --- once with the two constraints the implementation imposes, $\|\psi_{qj}\|_\infty \le M'$ and
$\|g_q\|_\infty \le M_g'$, and once with both switched off --- and counted how often the bound was
active. The two behave quite differently. The bound on the inner units is genuinely slack: the largest
fitted inner coefficient is $1.61$ along the knot schedule and $2.02$ over every configuration we fit,
against a clamp of $10$. The bound on the
outer units is not. It is active in $676$ of the $1120$ outer solves performed along the knot
schedule, and in $303$ of $336$ in a deliberately over-parameterized fit ($n=400$ with $k=26$); along
the schedule the unconstrained solve would have placed coefficients as large as $376$ (additive
target), $972$ (univariate) and $7680$ (multiplicative), against bounds between $5.7$ and $9.9$. The magnitude of the
\emph{fitted} coefficients does not reveal this, and reading it as though it did is a trap we fell
into ourselves: the bound is imposed in Lagrangian form, by raising a ridge parameter in factors of
ten until the coefficients comply, so once it engages it overshoots and the fitted coefficients settle
well below the bound --- the largest we observe is $8.84$ along the schedule and $9.31$ anywhere. What
has to be counted is the binding events, not the final magnitudes. (The bound itself is
$3\max_i |Y_i| + 1$ rather than a fixed $M_g'$, hence data-dependent: it is $5.7$--$6.0$ on the
additive and univariate targets and $9.5$--$9.9$ on the multiplicative one, whose response has a
wider range. The third constraint of \eqref{eq:sieve}, $\|g_q'\|_\infty \le L'$, is not imposed
explicitly; on a mesh of width $|\mathcal{I}|/k$ with bounded coefficients it holds automatically with
$L' = O(k)$, which is exactly the regime the proof of Lemma~\ref{lem:entropy} notes would cost a
further $\log k_n$ in the entropy. Since no sieve fit in this paper uses more than $41$ interior
knots, this affects no measurement reported here.)

What the constraints cost and what they buy are both worth stating. Along the knot schedule they cost
essentially nothing \emph{on the rate}, which is the claim that matters here: across the $63$ paired
refits the median ratio of unconstrained to constrained risk is $1.00$, and refitting the log--log slope
on the unconstrained column moves it by at most $0.11$: $-0.896$ to $-0.998$ (additive), $-0.855$ to
$-0.844$ (multiplicative), $-0.748$ to $-0.746$ (univariate), all at three replications. So none of
the reported convergence slopes is an artifact of the regularization. Individual configurations do
differ, and by more than that suggests: the largest single-fit ratio is $5.3\times$, and although
$12$ of the $15$ per-configuration medians at $n \ge 1600$ sit within $1.5\%$ of $1.00$, the three
exceptions run from $0.89$ (multiplicative, $n=1600$, where dropping the constraint \emph{helped}) to
$1.59$ (univariate, $n=1600$), with single fits there reaching $3.2$. The differences are scattered
rather than systematic, which is why they do not accumulate into a slope.

In the over-parameterized fit --- the regime the constraints exist for --- dropping them inflates the
risk by a median $18\times$ (additive) and $6\times$ (multiplicative). The textbook form of the
failure appears in one replication of the three --- the median one, so it is the $18\times$ just
quoted --- where the unconstrained fit reaches a \emph{lower} training loss ($1.7\times10^{-2}$
against $4.9\times10^{-2}$) and a risk $18\times$ worse; in the
other two the unregularized solve is worse on the training set as well, since escalating the ridge
conditions an ill-posed normal equation at the same time as it enforces the bound. Either way the
constrained fit is the one that wins: it has the smaller risk in every case and the smaller training
loss in five of the six. (These are \emph{paired} comparisons. Summarizing the two columns by their
medians separately, as we did at first, reverses the training-loss conclusion, because with three
replications the two medians come from different draws; when both fits see the same data the paired
ratio is the statistic to report.) The constraints are therefore load-bearing in both senses. They are what make $\mathcal{F}_n$ uniformly
bounded, which is what Lemmas~\ref{lem:entropy} and~\ref{lem:oracle} require, and they are active in
the fits.
\end{remark}

\begin{remark}[On the logarithmic factor]
\label{rem:log}
The classical spline rate recalled in Appendix~\ref{app:bspline} carries no logarithm, and it is worth
being precise about where ours enters. That classical result concerns an estimator that is
\emph{linear} in the spline coefficients: least squares over a $p$-dimensional linear space is an
orthogonal projection, its stochastic error is exactly $\sigma^2 p/n$ by a bias--variance identity,
and no covering argument is needed. The KAN sieve is parameterized by $p_n$ coefficients but is not a
linear space --- $g_q\circ T_q$ is a composition of two splines, which is not a spline --- so
projection is unavailable and the stochastic term must instead be controlled by chaining over a
covering of $\mathcal{F}_n$. A covering at resolution $\epsilon \asymp 1/n$ of a $p_n$-parameter
family costs $p_n\log n$, and that is the entire origin of the extra factor: it is the price of
replacing an exact projection identity by a uniform entropy bound, not a statement that KAN estimation
is intrinsically harder. Consistently with this, if the outer functions are held fixed --- for
instance $g_q = \mathrm{id}$, which reduces the model to a classical additive model --- then
$\mathcal{F}_n$ \emph{is} a linear space of dimension $O(k_n)$, the stochastic error returns to
$O(p_n/n)$ with no logarithm --- the projection identity above, requiring no covering and hence no
theorem beyond it --- and the clean rate $O(n^{-2r/(2r+1)})$ holds with $k_n \asymp n^{1/(2r+1)}$
\citep[Ch.~11]{gyorfi2002distribution}. Proposition~\ref{prop:nolog} gives a
condition under which the same is true for the full nonlinear sieve. We do not know whether the
logarithm can be removed unconditionally, and we prefer to state the rate we can prove.

Two things about that condition are worth spelling out. First, its localization is in $\|\cdot\|$ and
not in $\|\cdot\|_\infty$. Since $\|\cdot\| \le \|\cdot\|_\infty$ the $L^2$ ball is the larger of the
two, and it is the one the localization argument has to cover; a bound assumed only over the sup-norm
ball would not suffice. Second, the two halves of the bi-Lipschitz property play different roles. The
upper half --- $\|f_\theta - f_{\theta'}\|_\infty \le \kappa\|\theta-\theta'\|_\infty$ --- is already
proved, in Lemma~\ref{lem:entropy}, and is what supplies the covering. It is the lower half that is
missing: one needs $\|f_\theta - f_{\theta'}\| \gtrsim \mathrm{dist}(\theta,\theta')$ on the quotient
by the invariance group, which is exactly what turns an $L^2$ ball of radius $\delta$ back into a
parameter ball of radius $O(\delta)$ and so makes the entropy scale with $\log(\delta/\epsilon)$
rather than $\log(1/\epsilon)$. Proposition~\ref{prop:noident} shows the group is nontrivial, so this
is a statement about the geometry of the KAN parameterization modulo scaling, shifts and (when
$Q\ge2$) mixing --- not a routine verification. The same distinction explains why the penalty in
Theorem~\ref{thm:adaptive} must carry $\log n$: with only the \emph{global} entropy of
Lemma~\ref{lem:entropy} available, the local complexity that model selection charges for is
$p_{n,k}\log n$ and not $p_{n,k}$.
\end{remark}

\begin{remark}[Why Theorem~\ref{thm:kan_mult} is stated for hybrid targets]
\label{rem:hybrid}
Theorem~\ref{thm:kan_mult} is stated for genuinely hybrid targets --- each node independently additive
or multiplicative --- rather than for a purely multiplicative $f_0$. This matters for
Corollary~\ref{cor:minimax}: a target that mixes node types belongs to neither the purely additive nor
the purely multiplicative class, so a result covering only those two extremes would not cover
$\mathcal{F}^{\mathrm{KAN}}_r$. The multiplicative nodes enter only through the constant, via the
telescoping identity in Lemma~\ref{lem:approx} for a difference of $d$-fold products; the dependence
on $n$ --- and hence the rate --- is unaffected, so hybrid KANs attain the same exponent as purely
additive ones.
\end{remark}

\begin{remark}[The class must be a ball]
\label{rem:ball}
Restricting to a class with \emph{fixed} constants $(M,M_g,L,B)$ is essential and not a technicality.
The class of all functions admitting a KAN representation with univariate components in $W^r([0,1])$,
with no bound on their norms, is not totally bounded; its minimax risk is infinite for every $n$, and
no statement of the form $R_n \le C n^{-2r/(2r+1)}$ can hold over it. The upper bound of
Corollary~\ref{cor:minimax} is uniform over the ball precisely because the constant $C$ of
Theorem~\ref{thm:kan_add} depends on $f_0$ only through $(M,M_g,L,B)$.
\end{remark}

\begin{remark}[What the component bound does \emph{not} say]
\label{rem:components}
The final display of Corollary~\ref{cor:knot} is a statement about the deterministic best
approximation $\psi_{qj,k_n}$, not about the fitted component $\hat\psi_{qj}$. It is tempting to
conclude that $\mathbb{E}\|\hat\psi_{qj} - \psi_{qj}\|^2_{L^2} = O(n^{-2r/(2r+1)})$ as well, but that
is false in general: by Proposition~\ref{prop:noident} the map from components to $f$ is not
injective, so a sequence of fits with $\|\hat f_n - f_0\| \to 0$ may have $\hat\psi_{qj}$ that do not
converge to $\psi_{qj}$ at any rate.

The counterexample must be one the sieve admits, which rules out the most obvious candidate: the
scale group of Proposition~\ref{prop:noident}(a) with $a_n \to \infty$ is \emph{not} available inside
$\mathcal{F}_n$, because \eqref{eq:sieve} imposes $\|\psi_{qj}\|_\infty \le M'$ and
$\|g_q'\|_\infty \le L'$, which confine $a_q$ to a compact set. The sign flip does the job and needs
no room at all: taking $a_q = -1$ in Proposition~\ref{prop:noident}(a) --- that is, negating every
$\psi_{qj}$ of an additive node, or any one of them if the node is multiplicative, and replacing $g_q$
by $g_q(-\,\cdot\,)$ --- leaves $\|\psi_{qj}\|_\infty$, $\|g_q\|_\infty$, $\|g_q'\|_\infty$ and every
norm in Assumption~\ref{ass:target} exactly unchanged, the last because $\mathcal{I}$ is symmetric
about the origin. So both representations lie in $\mathcal{F}_n$ and both are centered. A sequence of exact minimizers may therefore alternate between them: $\hat f_n \to
f_0$ at the optimal rate while $\hat\psi_{qj}$ oscillates between $\psi_{qj}$ and $-\psi_{qj}$ and
converges to nothing. Consistent recovery of the components requires the normalizations and the
nondegeneracy of Proposition~\ref{prop:ident} --- which is why that proposition can only conclude
uniqueness \emph{up to a global sign} --- and is a genuinely different problem from estimating $f_0$.
\end{remark}

\begin{remark}[Very smooth targets]
\label{rem:rinfty}
The rule \eqref{eq:knot} is meant for a fixed finite $r$: as $r$ grows, $k_n \asymp n^{1/(2r+1)}$
grows more slowly, reflecting that smoother components need fewer knots. The formal limit
$r\to\infty$ (e.g.\ an analytic target such as $\sin$) should not be read as ``$k_n \to 1$ knot'';
rather, a fixed low-order spline space eventually suffices and the estimation error is dominated by
the parametric term $k_n/n$, so any slowly growing $k_n$ --- or a single spline space of moderate
fixed order --- attains a near-parametric rate. Two caveats. First, all of this is subject to
$r \le m$: with a fixed spline order the approximation error saturates at $k^{-m}$, so the exponent
one can actually realize is $\min\{r,m\}$, and the formal $r \to \infty$ limit is really the limit
$r \to m$ (Remark~\ref{rem:adaptive_cap}). Second, adaptation to an unknown $r$ in that range is
guaranteed by Theorem~\ref{thm:adaptive} and is handled empirically by cross-validation in
Section~\ref{sec:sim}.
\end{remark}

\begin{remark}[Independence across $i$, not just conditional moments]
\label{rem:indep}
Assumption~\ref{ass:noise} is stated for the \emph{pairs} $(X_i,\varepsilon_i)$ rather than as a list
of conditions on each $\varepsilon_i$ separately, and the difference matters. Conditional centering
and conditional sub-Gaussianity constrain the law of each $\varepsilon_i$ given $X_i$; they say
nothing about how the errors relate to one another. Take $\varepsilon_1 = \dots = \varepsilon_n = Z$
for a single bounded, mean-zero, $\sigma$-sub-Gaussian $Z$ drawn independently of the design. Every
$\varepsilon_i$ then satisfies both conditional requirements exactly, yet the sample carries a common
random shift: $n^{-1}\sum_i \varepsilon_i = Z$ does not vanish as $n$ grows, and
$n^{-1}\sum_i \varepsilon_i f(X_i) \to Z\,\mathbb{E}f(X)$ rather than $0$, uniformly over
$\mathcal{F}_n$. The empirical process $\{n^{-1}\sum_i \varepsilon_i (f-f_0)(X_i)\}$ that
Lemma~\ref{lem:oracle} controls therefore does not concentrate, and no rate follows --- the estimator
is fitting $f_0 + Z$, not $f_0$. Both results the lemma rests on
\citep[Ch.~11]{gyorfi2002distribution} and \citep[Thm.~9.1]{vandegeer2000empirical} are stated for an
i.i.d.\ sample, so this is a requirement of the tools and not an artifact of how we invoke them.

Independence across $i$ is all that is added: the errors may still depend on the design through their
conditional law, which is what makes the assumption weaker than additive homoscedastic noise, and
they enter the proofs only through the two conditional bounds of Assumption~\ref{ass:noise}.
\end{remark}

\begin{remark}[Why the spline order must be at least two]
\label{rem:order2}
The condition on the order is $m \ge \max\{r,2\}$, not $m \ge r$, and the second half is not
redundant when $r = 1$. A spline of order $m=1$ has degree $0$: it is piecewise constant and jumps at
its interior knots. Such a $g$ has $g' = 0$ almost everywhere, so it satisfies the third constraint
of \eqref{eq:sieve}, $\|g'\|_\infty \le L'$, \emph{vacuously} while being discontinuous --- two
points $u,v$ on opposite sides of a knot give $|g(u)-g(v)|$ of order $M_g'$ for $|u-v|$ arbitrarily
small. The constraint therefore stops doing the job it was imposed for. Concretely, the proof of
Lemma~\ref{lem:entropy} passes from $\|g_{\theta'_q}'\|_\infty \le L'$ to
$|g_{\theta'_q}(T_{\theta,q}) - g_{\theta'_q}(T_{\theta',q})| \le L'|T_{\theta,q} - T_{\theta',q}|$,
an inference that needs $g_{\theta'_q}$ absolutely continuous. At $m=1$ it is false, and the failure
is not a matter of a worse constant: as $\|\theta-\theta'\|_\infty \to 0$ the node inputs
$T_{\theta,q}$ and $T_{\theta',q}$ still straddle a knot of the fixed function $g_{\theta'_q}$ at some
$x$, so $\theta \mapsto f_\theta$ is not sup-norm \emph{continuous} on $\Theta_n$ and no Lipschitz
constant works, $k_n$-dependent or not. The entropy bound --- hence
Theorems~\ref{thm:kan_add} and~\ref{thm:kan_mult} --- is lost. For $m \ge 2$
the elements of $S_{k,m}$ are continuous piecewise polynomials of degree $\ge 1$, lying in
$W^{1,\infty}$, and $\|g'\|_\infty$ is a genuine Lipschitz constant.

The restriction costs nothing. It binds only in the case $r=m=1$, and there it merely says that a
target of smoothness $r=1$ should be fitted with piecewise linear rather than piecewise constant
splines; the approximation rate $k^{-r}$ of \eqref{eq:qi_error} is unaffected, since it needs only
$r \le m$. Practice is far from the boundary in any case: our simulations use $m=4$, as does the
reference KAN implementation of \citet{liu2024kan}.
\end{remark}

\begin{remark}[The spline order caps the adaptivity range]
\label{rem:adaptive_cap}
The restriction $r \le m$ in Theorem~\ref{thm:adaptive} is not an artifact of the proof. Selection
over $k$ adapts to the unknown smoothness only within the approximation power of the fixed spline
space: a spline of order $m$ reproduces polynomials of degree $m-1$, so its best approximation error
on a $W^{r,\infty}$ ball saturates at $O(k^{-m})$ once $r > m$, no matter how $k$ is chosen.
Consequently $\hat f_{n,\hat k}$ attains $O((\log n/n)^{2m/(2m+1)})$ --- but no better --- on smoother
classes. Adapting to $r$ beyond $m$ requires selecting over the order as well, i.e.\ a two-parameter
grid $(k,m) \in \mathcal{K}_n \times \{2,\dots,\lceil\log n\rceil\}$; the same model-selection
argument applies verbatim to that finite family, at the price of an additional $\log n$ in the
cardinality term, and we state the single-order version because it is the one our simulations
implement.
\end{remark}

\begin{remark}[Two design pitfalls in the simulation]
\label{rem:design}
Both are easy to fall into and both invalidate the experiment.
\emph{(i) Smoothness.} A piecewise polynomial built by joining $t^{r+1}$ and $(1-t)^{r+1}$ at $t=1/2$
--- an apparently natural ``exactly $r$ times differentiable'' target --- is symmetric about $1/2$ and
therefore has a jump in its \emph{first} derivative there, for every $r$. Its smoothness is $3/2$ in
the $L^2$-Sobolev scale, not $r$; measuring its spline approximation exponent gives $1.46$, not $2$
(dashed line in Figure~\ref{fig:approx}, left). Because the kink sits at the fixed location $t=1/2$,
the error also depends systematically on whether the uniform grid places a knot there --- which
happens exactly when $k$ is odd, since the interior knots sit at $i/(k+1)$. Measuring it, the odd-$k$
error is almost exactly half the even-$k$ error at its neighbours (the ratio is $0.52$ throughout
$16 \le k \le 32$), so the error oscillates by a factor of nearly two with the parity of $k$ rather
than decaying smoothly. The grid plotted in Figure~\ref{fig:approx} is even throughout, which is why
the oscillation is not visible there; fitting the exponent on odd and even $k$ together gives $1.45$
against the $1.43$ of the even-only grid, so the conclusion about the target's smoothness does not
depend on this. (The error does not vanish at odd $k$: a simple knot yields a $C^{m-2}$ join, whereas
reproducing a kink with cubics requires knot multiplicity $3$.)
\emph{(ii) Scale.} The phases $\phi_i$ in \eqref{eq:psi_target} do not affect smoothness --- Sobolev
membership depends only on $|a_i|$ --- but they matter for the experiment. With $\phi_i \equiv 0$ the
harmonics align at a common point, the sup-norm normalization $c_r$ shrinks the whole function, and
the resulting approximation error falls four to five orders of magnitude below the stochastic error
$\sigma^2 p_n/n$ at every $k$ the schedule selects: the ratio runs from $1.6\times10^{-5}$ at $n=400$
to $9.3\times10^{-5}$ at $n=25{,}600$, against $0.06$--$0.30$ for the randomized phases actually used.
The risk is then variance-dominated; the bias/variance balance the
theorems describe is never exercised; and since the variance term is
$\asymp k_n/n \asymp n^{-2r/(2r+1)}$ under the prescribed $k_n$, the fitted exponent reproduces the
prediction while testing nothing. We verified that this is what happens with $\phi_i \equiv 0$ and
chose randomized phases accordingly.
\end{remark}

\section{Proofs}
\label{app:proof}

Throughout, $c, C, C_1, \dots$ denote finite positive constants depending only on
$(Q,d,m,M,M_g,L,B,\sigma,c_X,C_X)$ and never on $n$, $k_n$ or $f_0$; their value may change from
line to line. Recall $\|g\|^2 = \int g^2 dP_X$ and $p_n = Q(d+1)(k_n+m)$.

The argument has three ingredients, isolated as Lemmas~\ref{lem:approx}--\ref{lem:oracle}: how well
the sieve approximates a KAN target, how large the sieve is in the sense of metric entropy, and what
least squares over a class of that size achieves. Theorems~\ref{thm:kan_add} and~\ref{thm:kan_mult}
follow by balancing the first against the third.

\subsection{Three lemmas}

\begin{lemma}[Approximation of KAN compositions]
\label{lem:approx}
Let Assumption~\ref{ass:target} hold and let the spline order satisfy $m \ge \max\{r,2\}$. There is a constant
$c_0 \ge 1$ depending only on $m$, and a constant $C_1$, such that for every $k \ge 1$ there exists
$f_k \in \mathcal{F}_n$ (the sieve \eqref{eq:sieve} built with $M' = c_0 M$, $M_g' = c_0M_g$,
$L' = c_0 L$ and $k_n = k$) with
\begin{equation}
\|f_k - f_0\| \ \le\ C_1 \max\{1, (M')^{d-1}\}\, k^{-r} .
\label{eq:approx_lemma}
\end{equation}
Moreover the univariate components of $f_k$ satisfy
$\|\psi_{qj,k} - \psi_{qj}\|_{L^2([0,1])} \le c B k^{-r}$ and
$\|g_{q,k} - g_q\|_{L^\infty(\mathcal{I})} \le c B k^{-r}$.
\end{lemma}

\begin{proof}
\emph{Step 0 (quasi-interpolants).} For an interval $\mathcal{J}$ let $\Pi_k : L^\infty(\mathcal{J})
\to S_{k,m}(\mathcal{J})$ be the de Boor--Fix quasi-interpolant. It is a bounded linear projector onto
$S_{k,m}$ and satisfies, for $1 \le r \le m$ and $p \in \{2,\infty\}$
\citep[Thm.~6.27]{schumaker2007spline},
\begin{equation}
\|h - \Pi_k h\|_{L^p(\mathcal{J})} \le c\, h_k^{\,r} \|h^{(r)}\|_{L^p(\mathcal{J})},
\qquad h_k = \frac{|\mathcal{J}|}{k+1},
\label{eq:qi_error}
\end{equation}
together with the stability bounds
\begin{equation}
\|\Pi_k h\|_{L^\infty} \le c_0 \|h\|_{L^\infty},
\qquad
\|(\Pi_k h)'\|_{L^\infty} \le c_0 \|h'\|_{L^\infty},
\label{eq:qi_stability}
\end{equation}
with $c, c_0$ depending only on $m$ \citep[Ch.~XII]{deboor2001splines}. Set
$\psi_{qj,k} = \Pi_k \psi_{qj}$ on $[0,1]$ and $g_{q,k} = \Pi_k g_q$ on $\mathcal{I}$.

For the inner functions the domain $[0,1]$ has unit length, so \eqref{eq:qi_error} with $p=2$ and
Assumption~\ref{ass:target}(i) give $\|\psi_{qj} - \psi_{qj,k}\|_{L^2} \le c(k+1)^{-r}B$ directly. For
the outer functions the domain is $\mathcal{I}$, whose length is \emph{not} bounded by a constant, and
\eqref{eq:qi_error} with $p=\infty$ reads
\begin{equation}
\begin{split}
\|g_q - g_{q,k}\|_{L^\infty(\mathcal{I})}
&\ \le\ c \left(\frac{|\mathcal{I}|}{k+1}\right)^{\!r} \big\|g_q^{(r)}\big\|_{L^\infty(\mathcal{I})}\\
&\ =\ \frac{c}{(k+1)^{r}}\; |\mathcal{I}|^{\,r}\big\|g_q^{(r)}\big\|_{L^\infty(\mathcal{I})}\\
&\ \le\ \frac{cB}{(k+1)^{r}} ,
\end{split}
\label{eq:outer_qi}
\end{equation}
where the last step is exactly the scale-free bound \eqref{eq:scalefree}. This is the one place where
that normalization is used, and it is not cosmetic: without \eqref{eq:scalefree} the squared risk
would carry an extra factor $|\mathcal{I}|^{2r}$ (Remark~\ref{rem:scalefree}). Together with
\eqref{eq:qi_stability} and Assumption~\ref{ass:target},
\begin{equation}
\begin{gathered}
\|\psi_{qj} - \psi_{qj,k}\|_{L^2([0,1])} \le cBk^{-r},\\
\|g_q - g_{q,k}\|_{L^\infty(\mathcal{I})} \le cBk^{-r},
\end{gathered}
\label{eq:comp_errors}
\end{equation}
and $\|\psi_{qj,k}\|_\infty \le c_0 M = M'$, $\|g_{q,k}\|_\infty \le c_0 M_g = M_g'$,
$\|g_{q,k}'\|_\infty \le c_0 L = L'$. Hence the function
$f_k(x) := \sum_{q} g_{q,k}(\widehat T_q(x))$, where $\widehat T_q$ is built from the
$\psi_{qj,k}$ with the same node types as $f_0$, lies in $\mathcal{F}_n$. Note the second bound in
\eqref{eq:comp_errors} is in the \emph{sup} norm; this is where Assumption~\ref{ass:target}(ii) is
used, and Remark~\ref{rem:outer_sup} explains why an $L^2$ bound would not suffice.

\emph{Step 1 (error in the node inputs).} Fix $q$. If the node is additive then
$|\widehat T_q(x) - T_q(x)| \le \sum_{j} |\psi_{qj,k} - \psi_{qj}|(x_j)$, so, writing $p_j$ for the
$j$-th marginal density of $P_X$ (bounded by $C_X$ under Assumption~\ref{ass:design}),
\begin{equation}
\begin{split}
\big\|\widehat T_q - T_q\big\|
&\ \le\ \sum_{j=1}^{d} \big\| (\psi_{qj,k}-\psi_{qj})(x_j) \big\|\\
&\ \le\ \sqrt{C_X} \sum_{j=1}^{d} \|\psi_{qj,k}-\psi_{qj}\|_{L^2([0,1])}\\
&\ \le\ \sqrt{C_X}\, d\, cB\, k^{-r}.
\end{split}
\label{eq:index_add}
\end{equation}
If the node is multiplicative, write $a_j = \psi_{qj}(x_j)$, $b_j = \psi_{qj,k}(x_j)$ and use the
telescoping identity
\begin{equation}
\prod_{j=1}^{d} a_j - \prod_{j=1}^{d} b_j
= \sum_{j=1}^{d} \Big(\prod_{i<j} b_i\Big)\,(a_j - b_j)\,\Big(\prod_{i>j} a_i\Big).
\label{eq:telescope}
\end{equation}
Since $|a_i| \le M \le M'$ and $|b_i| \le M'$, each summand is bounded pointwise by
$(M')^{d-1}|a_j - b_j|$, whence
\begin{equation}
\big\|\widehat T_q - T_q\big\|
\ \le\ (M')^{d-1} \sqrt{C_X}\, d\, cB\, k^{-r}.
\label{eq:index_mult}
\end{equation}
This is the step that produces the factor $(M')^{2(d-1)}$ in the squared risk. In both cases
$T_q$ and $\widehat T_q$ take values in $[-\bar M, \bar M] \subset \mathcal{I}$.

\emph{Step 2 (error after the outer function).} For each $q$ decompose
\begin{equation}
\begin{split}
g_{q,k}\big(\widehat T_q\big) - g_q\big(T_q\big)
={}&
\underbrace{\Big[g_{q,k}\big(\widehat T_q\big) - g_q\big(\widehat T_q\big)\Big]}_{\text{(I)}}\\
&+
\underbrace{\Big[g_q\big(\widehat T_q\big) - g_q\big(T_q\big)\Big]}_{\text{(II)}} .
\end{split}
\label{eq:outer_split}
\end{equation}
Term (I) is bounded pointwise by $\|g_{q,k} - g_q\|_{L^\infty(\mathcal{I})} \le cBk^{-r}$, using that
$\widehat T_q$ takes values in $\mathcal{I}$. Term (II) is bounded pointwise by
$L\,|\widehat T_q - T_q|$ because $\|g_q'\|_{L^\infty(\mathcal{I})} \le L$. Therefore
\begin{equation}
\big\| g_{q,k}(\widehat T_q) - g_q(T_q) \big\|
\ \le\ cBk^{-r} + L \big\|\widehat T_q - T_q\big\| ,
\end{equation}
and summing over $q = 1,\dots,Q$ and inserting \eqref{eq:index_add}--\eqref{eq:index_mult} gives
\eqref{eq:approx_lemma} with $C_1 = Q c B (1 + L d \sqrt{C_X})$.
\end{proof}

\begin{remark}[What the original decomposition omitted]
\label{rem:gap}
The decomposition \eqref{eq:outer_split} is essential: replacing $g_q$ by its spline approximation
contributes term (I), which is \emph{not} controlled by the Lipschitz bound on $g_q$. A bound of the
form $\|g_{q,k}(\widehat T_q) - g_q(T_q)\| \le L\|\widehat T_q - T_q\|$ would require
$g_{q,k} = g_q$ and is false otherwise. Term (I) is precisely what forces
Assumption~\ref{ass:target}(ii) to be stated in $W^{r,\infty}$ rather than $W^{r,2}$.
\end{remark}

\begin{lemma}[Metric entropy of the KAN sieve]
\label{lem:entropy}
There are constants $A < \infty$ and $D_m$ (depending only on $m$) such that for all
$0 < \epsilon \le A/2$,
\begin{equation}
\begin{gathered}
\log N\big(\epsilon, \mathcal{F}_n, \|\cdot\|_\infty\big)
\ \le\ p_n \log(A/\epsilon) + Q\log 2 ,\\
A = 4\kappa D_m \max\{M', M_g'\},
\end{gathered}
\end{equation}
where $\kappa = Q\big(1 + L' d \max\{1, (M')^{d-1}\}\big)$. In particular the bound does not involve
$k_n$ except through $p_n$.
\end{lemma}

\begin{proof}
Parameterize $f \in \mathcal{F}_n^\tau$ by the vector $\theta$ of all $Q(d+1)$ coefficient blocks,
each of length $k_n + m$. Because the B-spline basis is non-negative and forms a partition of unity,
$\|\sum_i c_i B_i\|_\infty \le \max_i |c_i|$; conversely the basis is $L^\infty$-stable,
$\max_i |c_i| \le D_m \|\sum_i c_i B_i\|_\infty$, with $D_m$ depending only on the order $m$
\citep[Ch.~XI]{deboor2001splines}. Hence the sup-norm constraints in \eqref{eq:sieve} confine
$\theta$ to the box $\Theta = [-D_m\max\{M',M_g'\},\, D_m\max\{M',M_g'\}]^{p_n}$. Write
$\Theta_n \subseteq \Theta$ for the set of parameters that additionally satisfy the constraints
$\|\psi_{qj}\|_\infty \le M'$, $\|g_q\|_\infty \le M_g'$, $\|g_q'\|_\infty \le L'$ of
\eqref{eq:sieve}, so that $\mathcal{F}_n^\tau = \{f_\theta : \theta \in \Theta_n\}$.

Next, the parameterization is Lipschitz \emph{on $\Theta_n$}. For $\theta, \theta' \in \Theta_n$ and
any $x$,
\begin{align}
\big|f_\theta(x) - f_{\theta'}(x)\big|
&\le \sum_{q=1}^{Q}
\Big[
\big|g_{\theta_q}(T_{\theta,q}) - g_{\theta'_q}(T_{\theta,q})\big| \notag \\
&\qquad\quad
+ \big|g_{\theta'_q}(T_{\theta,q}) - g_{\theta'_q}(T_{\theta',q})\big|
\Big] \notag \\
&\le \sum_{q=1}^{Q}
\Big[ \|\theta_q - \theta'_q\|_\infty + L' \big|T_{\theta,q} - T_{\theta',q}\big| \Big],
\end{align}
using the partition-of-unity bound for the first term and, for the second, that
$\theta' \in \Theta_n$ so $\|g_{\theta'_q}'\|_\infty \le L'$; both node inputs lie in $\mathcal{I}$
because $\theta,\theta' \in \Theta_n$ obey $\|\psi_{qj}\|_\infty \le M'$. (Two things are essential
here. The Lipschitz estimate is asserted only between two points of $\Theta_n$: a point of the
enclosing box $\Theta$ need not satisfy the derivative constraint, and for such a point the second
term is not controlled by $L'$. And the passage from $\|g_{\theta'_q}'\|_\infty \le L'$ to the
Lipschitz bound on $g_{\theta'_q}$ requires $g_{\theta'_q}$ to be absolutely continuous, which is why
the sieve is built with $m \ge 2$; at $m=1$ the implication fails outright, see
Remark~\ref{rem:order2}.) For an additive node
$|T_{\theta,q} - T_{\theta',q}| \le d \max_j \|\theta_{qj}-\theta'_{qj}\|_\infty$; for a
multiplicative node the telescoping identity \eqref{eq:telescope} gives the same bound with the extra
factor $(M')^{d-1}$. Hence $\|f_\theta - f_{\theta'}\|_\infty \le \kappa \|\theta - \theta'\|_\infty$
for $\theta,\theta' \in \Theta_n$, with $\kappa$ as stated. Note that $\kappa$ is free of $k_n$
\emph{because} the sieve imposes $\|g_q'\|_\infty \le L'$; without that constraint a spline with $k_n$
knots and bounded coefficients could have Lipschitz constant of order $k_n$, and an extra $\log k_n$
would appear.

It remains to cover $\Theta_n$, not $\Theta$. A $\delta/2$-net of $\Theta$ induces a $\delta$-net of
$\Theta_n$ \emph{whose points lie in $\Theta_n$}: discard the net points at distance more than
$\delta/2$ from $\Theta_n$ and replace each of the others by a point of $\Theta_n$ within $\delta/2$.
Applying this with $\delta = \epsilon/\kappa$ and using the standard volumetric bound for the box,
\begin{equation}
\begin{split}
N\big(\epsilon/\kappa, \Theta_n, \|\cdot\|_\infty\big)
&\ \le\ N\big(\epsilon/2\kappa, \Theta, \|\cdot\|_\infty\big)\\
&\ \le\ \Big(1 + \tfrac{2 D_m \max\{M',M_g'\}\,\kappa}{\epsilon}\Big)^{p_n}\\
&\ \le\ \Big(\tfrac{A}{\epsilon}\Big)^{p_n},
\end{split}
\end{equation}
the last step because $2 D_m \max\{M',M_g'\}\kappa = A/2$ and $1 \le A/(2\epsilon)$ on the stated
range $\epsilon \le A/2$. The image of such a net under
$\theta \mapsto f_\theta$ is an $\epsilon$-net of $\mathcal{F}_n^\tau$ by the Lipschitz bound, both
endpoints now being in $\Theta_n$. Taking the union over the $2^Q$ node-type patterns gives the
claim.
\end{proof}

\begin{lemma}[Oracle inequality for the sieve least-squares estimator]
\label{lem:oracle}
Let Assumptions~\ref{ass:design}--\ref{ass:noise} hold, let $\mathcal{F}_n$ be a class of functions
$[0,1]^d \to [-F,F]$ with $\log N(\epsilon,\mathcal{F}_n,\|\cdot\|_\infty) \le a_n\log(A/\epsilon)$,
and let $\hat f_n$ be the least-squares estimator over $\mathcal{F}_n$. Then for $n \ge 3$
\begin{equation}
\mathbb{E}\big[\|\hat f_n - f_0\|^2\big]
\ \le\
C \Big( \tfrac{a_n (\log n + \log A)}{n} + \inf_{f \in \mathcal{F}_n} \|f - f_0\|^2 \Big),
\label{eq:oracle}
\end{equation}
with $C$ depending only on $F$, $\sigma$ and $\|f_0\|_\infty$ --- in particular \emph{not} on $A$,
whose contribution is displayed.
\end{lemma}

\begin{proof}
This is the standard oracle inequality for least squares over a uniformly bounded class with
parametric-type metric entropy; see \citet[Ch.~11]{gyorfi2002distribution} for the version with a
bounded response and \citet[Thm.~9.1]{vandegeer2000empirical} for the sub-Gaussian errors of
Assumption~\ref{ass:noise}. Both are statements about an i.i.d.\ sample, which is why
Assumption~\ref{ass:noise} is imposed on the pairs $(X_i,\varepsilon_i)$ and not on the marginal
conditional laws of the $\varepsilon_i$ (Remark~\ref{rem:indep}). (The usual statement truncates the estimator at level $F$; here that step
is vacuous, since every $f \in \mathcal{F}_n$ already satisfies $\|f\|_\infty \le QM_g' = F$ by the
sup-norm constraints of \eqref{eq:sieve}, so the estimator analysed is exactly the empirical risk
minimizer \eqref{eq:erm} and not a modification of it.) The $\log n$ arises because the chaining bound is
evaluated at resolution $\epsilon \asymp 1/n$, giving
$\log N(1/n, \mathcal{F}_n, \|\cdot\|_\infty) \le a_n \log(An) = a_n(\log n + \log A)$, which is the
form displayed in \eqref{eq:oracle}.
\end{proof}

\subsection{Proof of Theorem \ref{thm:kan_add} and Theorem \ref{thm:kan_mult}}

\begin{proof}
We prove both at once; the additive case is the special case in which every node type is additive, and
then $\max\{1,(M')^{d-1}\}$ may be replaced by $1$ in \eqref{eq:approx_lemma}.

Every $f \in \mathcal{F}_n$ satisfies $\|f\|_\infty \le QM_g'=:F$, and $\|f_0\|_\infty \le QM_g \le F$.
By Lemma~\ref{lem:entropy} the entropy hypothesis of Lemma~\ref{lem:oracle} holds with
$a_n = p_n + Q$ and $A$ as in Lemma~\ref{lem:entropy}, so that $\log A \asymp d\log M'$ for a
multiplicative pattern. Since $A$ does not grow with $n$, $\log n + \log A \le C_A \log n$ for
$n \ge 3$ with $C_A = 1 + \log A/\log 3$, which we absorb into the constant below. Then
\eqref{eq:oracle} gives
\begin{equation}
\mathbb{E}\big[\|\hat f_n - f_0\|^2\big]
\ \le\
C\left( \frac{p_n \log n}{n} + \inf_{f\in\mathcal{F}_n}\|f - f_0\|^2 \right).
\end{equation}
By Lemma~\ref{lem:approx}, $\inf_{f\in\mathcal{F}_n}\|f-f_0\|^2 \le C_1^2\max\{1,(M')^{2(d-1)}\}k_n^{-2r}$,
and $p_n \asymp k_n$ by \eqref{eq:pn}. Hence
\begin{equation}
\mathbb{E}\big[\|\hat f_n - f_0\|^2\big]
\ \le\
C_2 \left( \frac{k_n \log n}{n} + k_n^{-2r} \right),
\label{eq:balance}
\end{equation}
with $C_2$ carrying the factor $\max\{1, (M')^{2(d-1)}\}$ in the hybrid case. The right-hand side is
minimized, up to constants, by equating the two terms:
$k_n \log n / n \asymp k_n^{-2r} \iff k_n^{2r+1} \asymp n/\log n$, i.e.\ by the choice
\eqref{eq:kn_log}. Substituting gives
\begin{equation}
\mathbb{E}\big[\|\hat f_n - f_0\|^2\big]
\ \le\
C_3 \left(\frac{\log n}{n}\right)^{\frac{2r}{2r+1}} .
\end{equation}
Every constant used depends on $f_0$ only through $(M,M_g,L,B)$, so the bound holds uniformly over
$\mathcal{F}^{\mathrm{KAN}}_r(M,M_g,L,B)$.
\end{proof}

\subsection{Proof of Proposition \ref{prop:nolog}}

\begin{proof}
Under \eqref{eq:local_entropy} the Dudley entropy integral of the localized class
$\mathcal{F}_n(\delta) = \{f \in \mathcal{F}_n : \|f - f_0\| \le \delta\}$ satisfies
\begin{equation}
\begin{aligned}
\mathcal{J}(\delta)
&:= \int_0^{\delta} \sqrt{\log N(\epsilon, \mathcal{F}_n(\delta), \|\cdot\|_\infty)}\, d\epsilon \\
&\ \le\ \sqrt{C_\star p_n} \int_0^{\delta} \sqrt{\log(\delta/\epsilon)}\, d\epsilon \\
&\ = \delta \sqrt{C_\star p_n} \int_0^1 \sqrt{\log(1/u)}\,du,
\end{aligned}
\end{equation}
and the last integral equals $\sqrt{\pi}/2$. So $\mathcal{J}(\delta) \le c\,\delta\sqrt{p_n}$, with no
logarithmic factor. By the standard localization theorem for least squares
\citep[Thm.~9.1]{vandegeer2000empirical} the rate $\delta_n$ of $\hat f_n$ towards the sieve optimum is
determined by the fixed-point relation $\sqrt{n}\,\delta_n^2 \gtrsim \mathcal{J}(\delta_n)$, which
here reads $\sqrt n \delta_n^2 \gtrsim \delta_n\sqrt{p_n}$, i.e.\ $\delta_n^2 \asymp p_n/n$. (That
theorem localizes in the empirical norm $\|\cdot\|_{L^2(P_n)}$; on the uniformly bounded class
$\mathcal{F}_n$ the two localizations differ by a constant factor outside an event of probability
$O(n^{-1})$, by the standard ratio-type inequality, which is why \eqref{eq:local_entropy} may be stated
for $\|\cdot\|$.) Combining
with Lemma~\ref{lem:approx} as in \eqref{eq:balance} gives
$\mathbb{E}\|\hat f_n - f_0\|^2 \lesssim k_n/n + k_n^{-2r}$, which is balanced by
$k_n \asymp n^{1/(2r+1)}$ and yields $O(n^{-2r/(2r+1)})$.
\end{proof}

\subsection{Proof of Proposition \ref{prop:noident}}

\begin{proof}
(a) Invariance of $f$ is immediate from
$g_q\big((a_q \sum_j \psi_{qj})/a_q\big) = g_q\big(\sum_j \psi_{qj}\big)$. For the constraints:
$\int_0^1 a_q\psi_{qj} = a_q\int_0^1 \psi_{qj} = 0$; and if $\tilde\mu_q$ denotes the law of
$a_q\sum_j\psi_{qj}(X_j)$ then
\begin{equation}
\begin{split}
\int g_q(u/a_q)\,d\tilde\mu_q(u)
&= \mathbb{E}\Big[g_q\Big(\tfrac{1}{a_q}\, a_q \textstyle\sum_j \psi_{qj}(X_j)\Big)\Big]\\
&= \mathbb{E}\Big[g_q\Big(\textstyle\sum_j \psi_{qj}(X_j)\Big)\Big]\\
&= \int g_q \, d\mu_q = 0 .
\end{split}
\end{equation}
So the whole one-parameter group $\{a_q \ne 0\}$ preserves the centered representation, and no
centering condition can remove it.

(b) With $d = 1$ write $c = \int_0^1 h(\psi(t))\,dt$, $\tilde\psi = h\circ\psi - c$ and
$\tilde g = g \circ h^{-1}(\cdot + c)$. Then
$\tilde g(\tilde\psi(x)) = g(h^{-1}(h(\psi(x)))) = g(\psi(x)) = f(x)$. Also
$\int_0^1\tilde\psi = 0$ by construction, and
$\int \tilde g \, d\tilde\mu = \mathbb{E}[\tilde g(\tilde\psi(X))] = \mathbb{E}[g(\psi(X))] = 0$.

(c) With $g_1 = g_2 = \mathrm{id}$ we have
$f(x) = \sum_j \psi_{1j}(x_j) + \sum_j \psi_{2j}(x_j) = \sum_j \phi_j(x_j)$, which depends on
$(\psi_{1j},\psi_{2j})$ only through $\phi_j = \psi_{1j}+\psi_{2j}$. Any decomposition
$\phi_j = \psi_{1j}' + \psi_{2j}'$ into centered summands yields the same $f$ and satisfies
\eqref{eq:centering}. The final shift statement is the computation
$g_q\big(\sum_j(\psi_{qj}+c_{qj}) - \sum_j c_{qj}\big) = g_q\big(\sum_j \psi_{qj}\big)$.
\end{proof}

\subsection{Proof of Proposition \ref{prop:ident}}

\begin{proof}
Let $T = \sum_j \psi_j$ and $\widetilde T = \sum_j \tilde\psi_j$. Since $g$ and $\tilde g$ are
continuous and strictly monotone on the relevant ranges, $h := \tilde g^{-1} \circ g$ is well defined,
continuous and strictly monotone on the range of $T$, and the hypothesis gives
\begin{equation}
\widetilde T(x) = h\big(T(x)\big) \qquad \text{for all } x \in [0,1]^d,
\label{eq:hT}
\end{equation}
the extension from a.e.\ $x$ to all $x$ being justified by continuity of both sides.

\emph{Step 1: each $\tilde\psi_j$ is a function of $\psi_j$.} Suppose $\psi_1(x_1) = \psi_1(x_1')$.
For any fixed $x_2,\dots,x_d$ the two points $(x_1,x_2,\dots)$ and $(x_1',x_2,\dots)$ give the same
value of $T$, hence by \eqref{eq:hT} the same value of $\widetilde T$, hence
$\tilde\psi_1(x_1) = \tilde\psi_1(x_1')$. So there is a function $A_1$ on the range $U$ of $\psi_1$
with $\tilde\psi_1 = A_1 \circ \psi_1$, and similarly $\tilde\psi_j = A_j \circ \psi_j$ for every $j$.
Each $A_j$ is continuous: $\psi_j : [0,1] \to U_j$ is a continuous surjection from a compact space
onto a Hausdorff one, hence a closed map and therefore a quotient map, so $A_j$ is continuous as soon
as $A_j \circ \psi_j = \tilde\psi_j$ is. This is what makes Step~2 applicable.

\emph{Step 2: a Pexider equation.} By hypothesis at least two of the $\psi_j$ are non-constant; relabel
so that these are $\psi_1, \psi_2$. Being continuous on $[0,1]$ and non-constant, their ranges $U, V$
are non-degenerate compact intervals. Fix values of $x_3,\dots,x_d$ and let
$s_0 = \sum_{j\ge3}\psi_j(x_j)$ and $c_0 = \sum_{j \ge 3} A_j(\psi_j(x_j))$. Then \eqref{eq:hT} reads
\begin{equation}
h(u + v + s_0) = A_1(u) + A_2(v) + c_0 \quad \text{for all } (u,v) \in U \times V .
\label{eq:pexider}
\end{equation}
This is Pexider's equation on a restricted rectangular domain. Since $h$ is continuous, the
classical theory of that equation \citep[\S3.1]{aczel1966lectures} --- in the restricted-domain form of
\citet{rado1987pexider} --- yields that $h$ is affine on the interval $U + V + s_0$, say
$h(w) = \alpha w + \beta$ there, and that $A_1, A_2$ are affine with the same slope $\alpha$.

\emph{Step 3: globalization and conclusion.} As $(x_3,\dots,x_d)$ varies, $s_0$ ranges over the
(connected) range of $\sum_{j \ge 3}\psi_j$ and the intervals $U + V + s_0$ overlap and cover the whole
range of $T$; on overlaps the affine representations agree, so $h(w) = \alpha w + \beta$ on the range
of $T$. (If $d = 2$, or if every $\psi_j$ with $j \ge 3$ is constant, $U+V+s_0$ is already the range
of $T$.) Since $h$ is strictly monotone, $\alpha \ne 0$. Now \eqref{eq:hT} becomes
\begin{equation}
\sum_{j=1}^d \tilde\psi_j(x_j) = \alpha \sum_{j=1}^d \psi_j(x_j) + \beta
\qquad \text{for all } x,
\end{equation}
and holding all coordinates but the $j$-th fixed shows $\tilde\psi_j - \alpha\psi_j$ is a constant
$b_j$, with $\sum_j b_j = \beta$. Finally $\tilde g = g \circ h^{-1}$ gives
$\tilde g(u) = g\big((u-\beta)/\alpha\big)$. Setting $a = \alpha$ completes the first claim.

If both representations are centered then $\int_0^1 \tilde\psi_j = \alpha\int_0^1\psi_j + b_j$ forces
$b_j = 0$ and hence $\beta = 0$; the scale normalization then gives
$1 = \sum_j\|\tilde\psi_j\|_{L^2}^2 = \alpha^2\sum_j\|\psi_j\|^2_{L^2} = \alpha^2$, so
$\alpha \in \{-1,+1\}$.
\end{proof}

\subsection{Proof of Corollary \ref{cor:minimax}}

\begin{proof}
\textbf{Upper bound.} Theorem~\ref{thm:kan_mult} applies to every node-type pattern and therefore to
all of $\mathcal{F}_r^{\mathrm{KAN}}(M,M_g,L,B)$, giving a single estimator $\hat f_n$ with
$\sup_{f_0} \mathbb{E}_{f_0}\|\hat f_n - f_0\|^2_{L^2} \le C(\log n/n)^{2r/(2r+1)}$; the infimum over
estimators is no larger. (It is worth noting that a result covering only the purely additive and the
purely multiplicative classes would \emph{not} suffice here, since a target mixing the two node types
lies in neither.)

\textbf{Lower bound.} Consider the univariate subclass
\begin{equation}
\begin{split}
\mathcal{W} = \big\{ f : [0,1]^d \to \mathbb{R} \ :\ &f(x) = h(x_1),\\
&h \in W^{r,2}([0,1]),\ \|h\|_\infty \le M,\\
&\|h^{(r)}\|_{L^2} \le B \big\} .
\end{split}
\end{equation}
We claim $\mathcal{W} \subset \mathcal{F}_r^{\mathrm{KAN}}(M,M_g,L,B)$. Given $h$, take the first node
additive with $\psi_{11} = h$ and $\psi_{1j} \equiv 0$ for $j \ge 2$, and $g_1 = \mathrm{id}$ on
$\mathcal{I}$; take $g_q \equiv 0$ and $\psi_{qj}\equiv 0$ for $q \ge 2$. Then
$f(x) = h(x_1)$, and the constraints of Assumption~\ref{ass:target} hold because
$\|\mathrm{id}\|_{L^\infty(\mathcal{I})} = \bar M + 1 \le M_g$, $\|\mathrm{id}'\|_\infty = 1 \le L$,
and the scale-free bound \eqref{eq:scalefree} reads
$|\mathcal{I}|^{\,r}\|\mathrm{id}^{(r)}\|_\infty = |\mathcal{I}|$ if $r=1$ and $0$ if $r \ge 2$, in
either case at most $B$ (this is where the standing conditions $M_g \ge \bar M + 1$, $L \ge 1$ and
$B \ge |\mathcal{I}|$ are used; the last is the only place the normalization of
Remark~\ref{rem:scalefree} costs anything, and it costs a constant that does not depend on $n$, so
the rate statement is unaffected).

Since the response depends on $x_1$ only and $X_1$ is uniform on $[0,1]$, estimating $f \in
\mathcal{W}$ is exactly univariate nonparametric regression over a Sobolev ball of smoothness $r$
with Gaussian errors, for which the classical minimax lower bound
\citep[Thm.~2.9 and \S2.6]{tsybakov2008nonparametric} gives
$R_n(\mathcal{W}) \ge c\, n^{-2r/(2r+1)}$ for some $c>0$ and all large $n$. Minimax risk is monotone
under inclusion of the parameter class, so
$R_n(\mathcal{F}_r^{\mathrm{KAN}}) \ge R_n(\mathcal{W}) \ge c\,n^{-2r/(2r+1)}$.
\end{proof}

\subsection{Proof of Corollary \ref{cor:knot}}

\begin{proof}
The decomposition \eqref{eq:tradeoff_main} is \eqref{eq:balance}, established in the proof of
Theorem~\ref{thm:kan_add}. The function $k \mapsto k^{-2r} + k\log n/n$ is, up to constants,
minimized where its two terms agree, which happens at $k^{2r+1} \asymp n/\log n$; substituting gives
the stated rate. The final display of Corollary~\ref{cor:knot} is the first bound of \eqref{eq:comp_errors} in
Lemma~\ref{lem:approx} with $k = k_n$: it concerns the deterministic best spline approximation
$\psi_{qj,k_n}$ to $\psi_{qj}$, and involves no estimation. As stressed in
Remark~\ref{rem:components}, it does not transfer to the fitted components $\hat\psi_{qj}$, which by
Proposition~\ref{prop:noident} need not converge to $\psi_{qj}$ at all.
\end{proof}

\subsection{Proof of Theorem \ref{thm:adaptive}}

\begin{proof}
The collection $\{\mathcal{F}_{n,k}\}_{k\in\mathcal{K}_n}$ is a finite family of models, of cardinality
$|\mathcal{K}_n| \le 1 + \log_2 n$. Each $\mathcal{F}_{n,k}$ is uniformly bounded (by $QM_g'$) and, by
Lemma~\ref{lem:entropy}, has metric entropy $\log N(\epsilon,\mathcal{F}_{n,k},\|\cdot\|_\infty) \le
p_{n,k}\log(A/\epsilon) + Q\log 2$ with $p_{n,k}\asymp k$. Assign to model $k$ the weight
$w_k = 2^{-p_{n,k}}$, so that $\sum_{k\in\mathcal{K}_n} w_k \le 1$. The model-selection theorem for
penalized least squares with sub-Gaussian noise \citep[Thm.~8.1]{massart2007concentration} requires a
penalty dominating $\kappa'\sigma^2\big(\mathcal{D}_{n,k} + \log(1/w_k) + \log|\mathcal{K}_n|\big)/n$,
where $\mathcal{D}_{n,k}$ is \emph{not} the parameter count but the local complexity of
$\mathcal{F}_{n,k}$: the fixed point of its localized Dudley integral. This distinction is the whole
content of the logarithm. For a \emph{linear} model $\mathcal{D}_{n,k} = p_{n,k}$ and the penalty
$\sigma^2 p_{n,k}/n$ suffices; for our nonlinear sieve the only entropy bound available is the
\emph{global} one of Lemma~\ref{lem:entropy}, whose Dudley integral evaluated at resolution
$\epsilon \asymp 1/n$ gives $\mathcal{D}_{n,k} \asymp p_{n,k}\log n$ --- the same factor, arising the
same way, as in Lemma~\ref{lem:oracle}. Since $\log(1/w_k) = p_{n,k}\log 2$ and
$\log|\mathcal{K}_n| \le \log(1+\log_2 n)$ are both dominated by $\mathcal{D}_{n,k}$, the penalty
$\mathrm{pen}(k) = \kappa\sigma^2 p_{n,k}\log n / n$ is admissible; see also
\citet[\S4]{barron1999risk}, which treats families of models specified by a metric entropy of exactly
this form, and \citet[Ch.~12]{gyorfi2002distribution}. The theorem gives, for $\kappa$ large enough,
\begin{equation}
\begin{split}
\mathbb{E}\!\left[\|\hat f_{n,\hat k} - f_0\|^2\right]
&\ \le\ \frac{C\sigma^2}{n}
\ +\ C \min_{k\in\mathcal{K}_n}\\
&\Big\{ \inf_{f\in\mathcal{F}_{n,k}} \|f - f_0\|^2 + \tfrac{p_{n,k}\log n}{n} \Big\},
\end{split}
\label{eq:oracle_select}
\end{equation}
with $C$ depending only on the constants of Assumptions~\ref{ass:design}--\ref{ass:target} and not on
$r$. Fix any $r\ge 1$ and $f_0\in\mathcal{F}^{\mathrm{KAN}}_r$. By Lemma~\ref{lem:approx},
$\inf_{f\in\mathcal{F}_{n,k}}\|f-f_0\|^2 \le C_1' k^{-2r}$ for every $k$ --- this is where $r \le m$
enters, since Lemma~\ref{lem:approx} requires the spline order to be at least the smoothness, and
without it the bound saturates at $k^{-2m}$ (Remark~\ref{rem:adaptive_cap}) --- so the bracketed term in
\eqref{eq:oracle_select} is at most $C_1' k^{-2r} + C_2' k\log n/n$. Let $k^\star$ be the smallest
element of the dyadic grid $\mathcal{K}_n$ that is $\ge (n/\log n)^{1/(2r+1)}$; since consecutive grid
points differ by a factor $2$, $k^\star \le 2 (n/\log n)^{1/(2r+1)}$, and evaluating the bracket at
$k=k^\star$ gives $O\big((\log n/n)^{2r/(2r+1)}\big)$. (Such a $k^\star \in \mathcal{K}_n$ exists
because $(n/\log n)^{1/(2r+1)} \le n$ for $r\ge 1$.) As the minimum in \eqref{eq:oracle_select} is no
larger than this value, and $\sigma^2/n$ is of smaller order, the claim follows. Neither the estimator
nor the penalty uses $r$, so the bound holds simultaneously for every $r \in [1,m]$ --- the upper
endpoint being exactly the restriction imposed above by Lemma~\ref{lem:approx}, beyond which the
approximation term saturates at $k^{-2m}$ (Remark~\ref{rem:adaptive_cap}).
\end{proof}

\section{Further Experimental Details}
\label{app:experiments}

\paragraph{The smoothness of the target is exactly $r$.}
Membership $\psi \in W^{s}([0,1])$ is equivalent to $\sum_i (1+i^2)^{s}a_i^2 < \infty$. Substituting
the coefficients $a_i = i^{-(r+1/2)}/\log(i+1)$ of \eqref{eq:psi_target} gives a summand
$\asymp i^{2(s-r)-1}/\log^2 i$, which converges at $s=r$ by Cauchy condensation (the condensed summand
is $\asymp i^{-2}$) and diverges for every $s > r$. Hence $\psi_r \in W^r([0,1])$ and
$\psi_r \notin W^{s}$ for $s > r$: its smoothness is exactly $r$, which is what makes it usable for
testing an exponent.

\paragraph{The over-specified fit in Table~\ref{tab:slopes}.}
One row of Table~\ref{tab:slopes} reports a \emph{hybrid} KAN --- one additive and one multiplicative
node --- fitted to the purely additive target. It is over-specified rather than misspecified: the
truth is recovered by zeroing the second outer function, so the class still contains $f_0$ and the
predicted exponent is unchanged. What the redundancy costs is a factor $5.6$ in the constant (median
over $n$, range $2.5$--$7.6$) and a steeper fitted exponent, $-1.08$ against the matched $-0.86$ ---
the redundant node is the high-variance row of Table~\ref{tab:slopes}, with much the widest interval. This is the closest the experiments come to the situation the theory actually assumes, where
the sieve is the union $\bigcup_\tau \mathcal{F}_n^\tau$ over node-type patterns and the estimator is
not told which pattern generated the data.

\paragraph{Why the P3 slopes should not be read quantitatively.}
The fitted slopes $-0.81,\,-0.86,\,-1.01$ for $r=1,2,3$ are all steeper than the corresponding
upper-bound exponents $-0.67,-0.80,-0.86$, for the same reason as in P2: a steeper slope is
permitted by an upper bound, not predicted by it. The mechanism is worth stating precisely,
because it also governs the noise and dimension sweeps. The estimation term of
Corollary~\ref{cor:knot} is $\sigma^2 p_n/n$ with $p_n = Q(d+1)(k_n+m)$, and for large $r$ the knot
count barely moves: at $r=3$ it runs only over $k_n \in \{5,\dots,9\}$, so $k_n$ grows by a factor
$1.8$ across the sweep while $p_n$, damped by the fixed offset $m$, grows by only $1.44$. The
estimation term therefore decays at $-0.915$, not at the $-0.865$ of $k_n/n$; and because the
approximation term is largely spent by $n \approx 3200$ --- the measured risk exceeds
$\sigma^2 p_n/n$ by $70\%$ at $n=400$, and from $n=3200$ on departs from it by between $+1\%$ and
$+21\%$, non-monotonically because $k_n$ is a step function of $n$: it takes the value $8$ at both
$n=6400$ and $12{,}800$, halving the estimation term while the approximation term is held
fixed --- the fitted slope steepens further, to
the measured $-1.01$. Nothing here is a statement about the exponent: it is the mixture of the two terms of
\eqref{eq:tradeoff_main} changing composition over the range of $n$ that is accessible. The predicted
slopes for $r=1,2,3$ are in any case too close together to be resolved at these sample sizes, which is
why the clean quantitative test of the $r$-dependence is the approximation experiment P1, whose
measured exponents $1.09, 1.96, 2.88$ match $r=1,2,3$ almost exactly.

\subsection{Additional Baselines}
\label{app:baselines}

\begin{figure}[t]
    \centering
    \includegraphics[width=0.82\columnwidth]{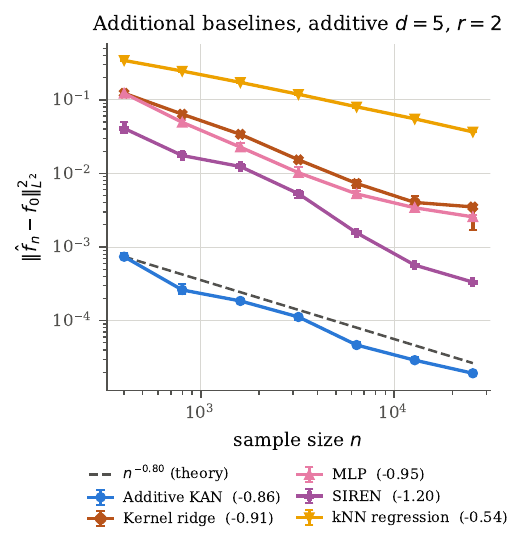}
    \caption{\textbf{All baselines on the additive target}, $d=5$, $r=2$. The matched KAN is an order
    of magnitude below every structure-free or generic-architecture competitor at each $n$, while the
    fitted slopes (in parentheses) show no rate deficit for the KAN. Every estimator here is fitted on
    the same sample sizes; only the \emph{dimension} sweep of Figure~\ref{fig:dimslopes} truncates
    kernel ridge, for the reason given there.}
    \label{fig:baselines}
\end{figure}

Reviewers of an earlier version of this work asked for baselines beyond the ReLU MLP, naming
DeepONet, SIREN and kernel regression. DeepONet is an operator-learning architecture whose input is a
discretized \emph{function}, so it does not apply to estimating a scalar $f:[0,1]^d \to \mathbb{R}$
from point samples and we do not run it. The other two do apply, and each carries a prediction. Figure~\ref{fig:baselines} collects them
alongside the estimators already discussed.

\paragraph{Kernel ridge regression.}
Kernel ridge with an RBF kernel is a \emph{global} smoother that makes no structural assumption on
$f$, so it plays the same theoretical role as $k$-NN --- minimax over $W^r([0,1]^d)$ only at
$-2r/(2r+d)$ --- while removing local averaging as an alternative explanation of the gap. Bandwidth
and ridge parameter are chosen by exact leave-one-out cross-validation, evaluated in closed form from
one eigendecomposition per bandwidth. The dual solve is $O(n^3)$ in time and $O(n^2)$ in memory --- the
kernel matrix reaches $5.2$\,GB at the largest sample size --- so it is performed in place; this is a
cost, not a restriction, and the baseline is run at every $n$ the KAN is.

On the additive target with $d=5$ its risk is a median $166\times$ that of the matched KAN (range
$136$--$244$ across $n$), comparable to the MLP's $123\times$ and far below $k$-NN's $\approx 10^3$.
Its fitted slope, $-0.91$, does \emph{not} exhibit the curse: at $d=5$ the structure-free rate $-0.44$
is simply not yet visible at these sample sizes, and the KAN ($-0.86$) and MLP ($-0.95$) are equally
steep over the same range. The prediction bites in the dimension sweep instead.

\begin{figure}[t]
    \centering
    \includegraphics[width=0.85\columnwidth]{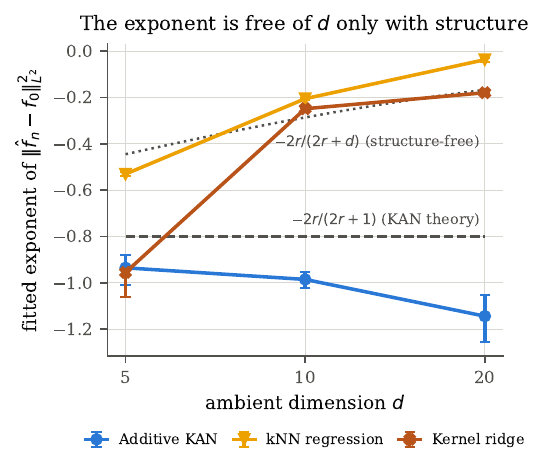}
    \caption{\textbf{Fitted exponent against ambient dimension.} The two reference curves apply to
    different estimators and are not alternatives: the KAN is predicted to sit on the flat line
    $-2r/(2r+1)$, and only the structure-free baselines are predicted to follow $-2r/(2r+d)$. The
    additive-KAN exponent is flat in $d$; the two baselines degrade toward the curse curve ($k$-NN past
    it), kernel ridge falling from indistinguishable-from-the-KAN at $d=5$ onto it by $d=20$. All three estimators are
    fitted on the same sample sizes, $n \le 12{,}800$: the $n^3$ kernel solve at $n=25{,}600$ costs
    about a minute per fit and runs one at a time, which is another half hour across three dimensions
    and eight replications for a slope six sample sizes already determine. Error bars are $95\%$
    bootstrap intervals.}
    \label{fig:dimslopes}
\end{figure}

Figure~\ref{fig:dimslopes} is the sharpest single picture of the paper's central claim. The two
predictions it draws are different curves and apply to different estimators: the KAN is predicted to
sit on the flat line $-2r/(2r+1)$, and only the structure-free baselines are predicted to follow
$-2r/(2r+d)$. Read that way, the kernel baseline is the informative one: at $d=5$ it is
indistinguishable from the KAN ($-0.95$ against $-0.94$), and by $d=20$ it has lost most of its
exponent and sits on the curse-of-dimensionality curve ($-0.179$ against the predicted $-0.167$;
$-0.248$ against $-0.286$ at $d=10$), while the KAN exponent does not drift
($-0.94, -0.99, -1.14$; these differ slightly from the values quoted in the main text because all
three estimators are refitted here on the shorter common range, not because the fits differ). That an
estimator can look as
good as the KAN in five dimensions and collapse by twenty is exactly the content of the claim that the
KAN exponent is free of $d$ \emph{because the target has KAN structure}. ($k$-NN is if anything worse
than the curse predicts at $d=10,20$; that is a finite-sample property of local averaging in high
dimension, not a contradiction of a rate that only the minimax-optimal estimator is required to
attain.)

\paragraph{SIREN.}
A sine-activation network carries no separate rate theory, so the prediction for it is the same as for
the ReLU MLP: a comparable exponent, with the difference appearing in the constant. Its frequency
scale $w_0$ has to be chosen with some care, and it is worth recording why. The value $w_0 = 30$
recommended by \citet{sitzmann2020implicit} is calibrated for fitting images and audio, whose content
is high-frequency; our targets have Sobolev smoothness $r \le 3$. Sweeping $w_0$ over
$\{1,3,10,30\}$ at $n=6400$ gives median test errors $1.7\times10^{-3}$, $1.6\times10^{-3}$,
$7.8\times10^{-3}$ and $1.01$, so we use $w_0=3$. The failure mode changes along the sweep, and the
training losses say how: at $w_0=10$ the network interpolates (training MSE $5\times10^{-10}$) and
generalizes five times worse than at $w_0=3$; at $w_0=30$ it does not fit at all, reaching a training
MSE of $0.85$ and a test error of $1.01$ against a unit-variance target --- no better than predicting
the mean. At the smallest sample size, $n=400$, every $w_0 \ge 3$ interpolates and the two largest
return test errors above $1$. Reporting the $w_0=30$ number as a baseline would have been a strawman.

So tuned, SIREN attains a fitted slope of $-1.20$ on the additive target, with a risk a median
$47\times$ that of the matched KAN. Like the MLP it shows no rate deficit --- its slope is if anything
steeper than the KAN's --- and the separation is again one of constants, which is the reading the
theory supports. Its ratio to the KAN does narrow across the sweep (rising from $55\times$ to
$67\times$ over the three smallest sample sizes, then falling to $17\times$ at the largest), so unlike
the MLP its gap is not flat in $n$; with a fixed training budget
we do not read that as a rate statement, for the reason given in the main text.

\paragraph{How much of the risk is approximation error.}
The convergence panels report the total risk. To separate its two parts we refit the same sieve, at
the same $k$, to a large \emph{noiseless} sample ($n = 40{,}000$, three seeds); what remains is the
approximation error of the sieve at that $k$. This is a direct check on Lemma~\ref{lem:approx} that a
convergence slope cannot give, and it says which regime each panel is in.

Both panels carry a substantial bias, the multiplicative one more so. Along the knot schedule the
ratio of total risk to approximation error has median $1.5$ on the multiplicative target (range
$1.2$--$1.9$ across $n$) and $2.7$ on the additive one (range $2.1$--$7.3$): the multiplicative panel
is close to pure approximation error, the additive panel is a genuine mixture. Fitted against $n$, the
approximation error along the schedule decays at $-0.80$ (multiplicative) and $-0.64$ (additive),
against measured risk slopes of $-0.85$ and $-0.86$.

Read against $k$ rather than $n$, the additive panel gives the cleaner check on Lemma~\ref{lem:approx}:
its approximation error decays like $k^{-3.82}$ against the predicted $k^{-2r}=k^{-4}$, and the fitted
exponent is identical across all three seeds. The multiplicative panel decays faster than predicted,
around $k^{-4.7}$, but that number should not be read precisely: the fit there is non-convex and the
error is markedly non-monotone in $k$ ($2.2\times10^{-4}$ at $k=8$, rising to $2.6\times10^{-4}$ at
$k=9$, then falling to $4.5\times10^{-5}$ at $k=10$), so a power law through seven such points is
unstable --- across seeds the fitted exponent moves over $[-5.03,-4.64]$. The non-monotonicity itself is stable
(the minimum and median over seeds agree at every $k$ to within a few percent), so it reflects how the
uniform knot grid aligns with the target rather than an artifact of the optimizer, and it is what
produces the visible steps in the multiplicative convergence curve. Those steps are a finite-$k$
effect, not a departure from the predicted rate.

\paragraph{Why the fitted slopes drift with $d$ and $\sigma$.}
The fitted slopes are not constant across the dimension and noise sweeps: they steepen monotonically,
from $-0.88$ to $-1.08$ as $d$ goes from $5$ to $20$, and from $-0.88$ to $-1.01$ as $\sigma$ goes from
$0.05$ to $0.5$. Read naively this contradicts the claim that the exponent depends on neither. It does
not, and the reason is worth giving quantitatively, because it is a prediction of
Corollary~\ref{cor:knot} rather than an excuse. A fitted log--log slope over a finite range of $n$ is
not an exponent: it is the slope of the \emph{sum} of the two terms in \eqref{eq:tradeoff_main}, and
anything that changes their relative size moves it. Both terms are measurable here. For the
approximation term we use the noiseless refits above, extended to $d \in \{10,20\}$. For the
estimation term we use $\sigma^2 p_n/(n-p_n)$, the classical finite-sample form: the asymptotic
$\sigma^2 p_n/n$ is inadequate in the corner of this design where $p_n/n$ reaches $0.58$
($d=20$, $n=400$).

With both terms measured and nothing fitted, the sum reproduces what is observed. Across the dimension
sweep the predicted drift in the fitted slope is $-0.177$ against a measured $-0.195$; across the noise
sweep it is $-0.079$ against a measured $-0.125$. The individual slopes agree to within $0.10$
throughout, the measured value being the steeper in all seven --- the model captures the direction and
most of the size of the drift, and understates the rest. The mechanism is the
same in both sweeps: raising $d$ raises $p_n = Q(d+1)(k_n+m)$ and hence the estimation term, and
raising $\sigma$ scales that term by $\sigma^2$, while the approximation term is untouched in either
case; the mixture therefore shifts toward its estimation term, whose slope is the steeper of the two,
and the fitted slope follows. The approximation term's share falls from $0.25$ to $0.05$ at $n=400$ as
$d$ goes from $5$ to $20$, and from $0.25$ to essentially zero as $\sigma$ goes from $0.05$ to $0.5$.

Two further readings follow. First, the measured approximation error is essentially \emph{flat} in $d$
--- relative to $d=5$ it is $1.06$ at $d=10$ and $1.08$ at $d=20$ --- whereas Lemma~\ref{lem:approx}
bounds the node-input error by a sum of $d$ univariate errors and so predicts $\sqrt{d}$ growth
($1.41$ and $2.00$) for this target, whose components are $\psi_r/\sqrt{d}$. The bound is loose here
because the $d$ coordinate errors are independent and nearly mean-zero, so they cancel rather than
accumulate; being an upper bound, it is entitled to be. This is also why the constant grows more slowly
than $p_n \propto d+1$ would suggest: at $n=25{,}600$ the risk is $1.40$ and $2.63$ times its $d=5$
value, against $1.83$ and $3.50$. Second, the estimation term that fits the data carries no logarithm.
Corollary~\ref{cor:knot} allows $p_n\log n/n$, which over this sweep is $\log n \in [6,10]$
times larger; the measured residual tracks $\sigma^2 p_n/(n-p_n)$ instead. That is consistent with the
upper bound, and it is what Proposition~\ref{prop:nolog} would predict, but a fitted decomposition over
seven sample sizes is evidence about this estimator on these targets and not a proof.

\begin{figure}[t]
    \centering
    \includegraphics[width=\columnwidth]{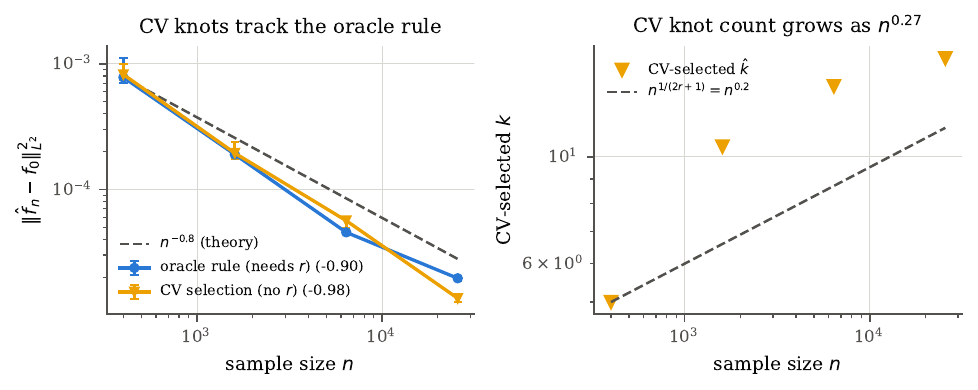}
    \caption{\textbf{Adapting to unknown smoothness.} The knot rule of Corollary~\ref{cor:knot} uses
    $r$; here $k$ is chosen instead by $3$-fold cross-validation, which uses no knowledge of $r$.
    \textbf{Left:} the CV-selected estimator tracks the oracle rule --- never more than $24\%$ worse,
    and $31\%$ better at the largest $n$. The two fitted slopes ($-0.98$ against $-0.90$) come from
    four points and are not evidence that the exponents agree.
    \textbf{Right:} the CV-selected knot count grows with $n$ (fitted $0.27$ against the theoretical
    $1/(2r+1)=0.2$; four sample sizes and integer $k$). Additive target, $d=5$, $r=2$.}
    \label{fig:adaptive}
\end{figure}

\begin{figure}[t]
    \centering
    \includegraphics[width=0.485\columnwidth]{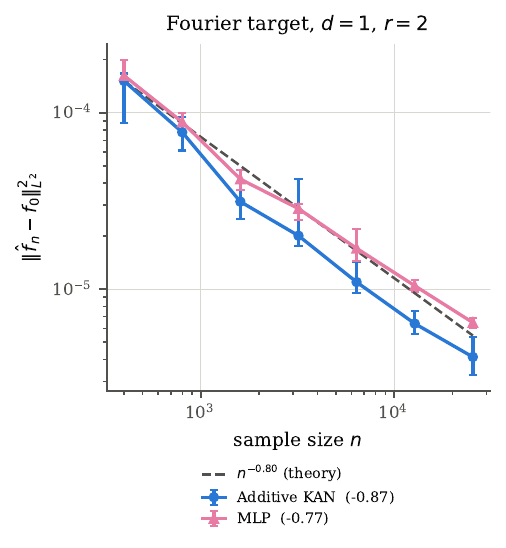}
    \hfill
    \includegraphics[width=0.485\columnwidth]{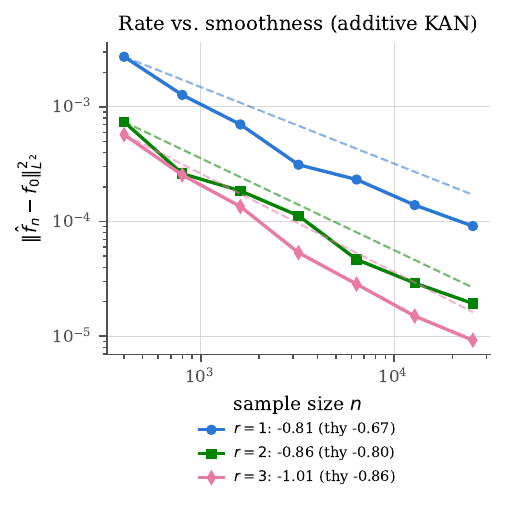}
    \caption{\textbf{Left:} the univariate Fourier target, $d=1$, $r=2$ (Theorem~\ref{thm:kan_add}).
    \textbf{Right:} dependence of the exponent on smoothness (P3); dashed lines are the predicted
    slopes $-2r/(2r+1)$ for $r=1,2,3$. As discussed in the main text, the predicted slopes are too
    close together to be resolved at these sample sizes; the clean $r$-dependence is the approximation
    experiment of Figure~\ref{fig:approx} (left).}
    \label{fig:smoothness}
\end{figure}

\begin{table*}[t]
\centering
\caption{Fitted log--log slopes as in Table~\ref{tab:slopes}, with the exponent predicted by the
theory in the last column. \emph{Theory} is $-2r/(2r+1)$ for each KAN whose class contains the target
and the structure-free rate $-2r/(2r+d)$ for $k$-NN. It is left blank for the MLP because the rates cited for
deep networks apply to capacity-controlled \emph{sequences} of networks, not to one architecture at
one budget, so they predict no exponent for the network we fit.}
\label{tab:slopes_full}
\footnotesize
\begin{tabular}{llrrr}
\toprule
Target & Estimator & Fitted slope & 95\% CI & Theory \\
\midrule
additive $d=5$ & Additive KAN & $-0.861$ & $[-0.918, -0.828]$ & $-0.800$ \\
additive $d=5$ & Hybrid KAN & $-1.077$ & $[-1.183, -0.944]$ & $-0.800$ \\
additive $d=5$ & MLP & $-0.948$ & $[-0.967, -0.925]$ & -- \\
additive $d=5$ & kNN regression & $-0.539$ & $[-0.546, -0.532]$ & $-0.444$ \\
Fourier $d=1$ & Additive KAN & $-0.869$ & $[-0.937, -0.761]$ & $-0.800$ \\
Fourier $d=1$ & MLP & $-0.766$ & $[-0.808, -0.733]$ & -- \\
mult.\ $d=2$ & Multiplicative KAN & $-0.848$ & $[-0.890, -0.526]$ & $-0.800$ \\
mult.\ $d=2$ & MLP & $-0.684$ & $[-0.698, -0.673]$ & -- \\
additive, $r=1$ & Additive KAN & $-0.811$ & $[-0.896, -0.792]$ & $-0.667$ \\
additive, $r=2$ & Additive KAN & $-0.861$ & $[-0.918, -0.828]$ & $-0.800$ \\
additive, $r=3$ & Additive KAN & $-1.010$ & $[-1.038, -0.963]$ & $-0.857$ \\
\bottomrule
\end{tabular}
\end{table*}

\begin{figure}[t]
    \centering
    \includegraphics[width=0.85\columnwidth]{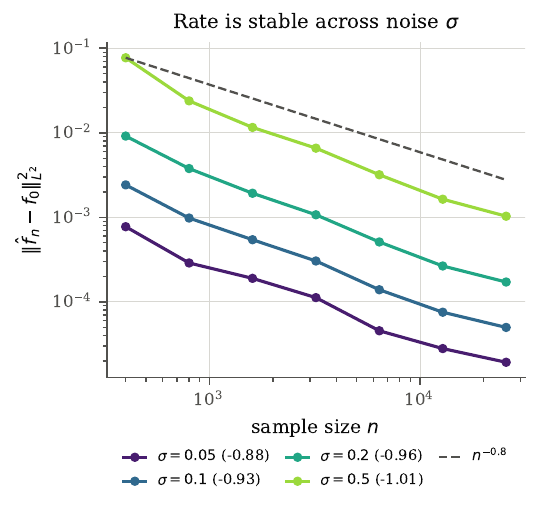}
    \caption{\textbf{Robustness to noise.} Additive target, $d=5$, $r=2$, with
    $\sigma \in \{0.05,0.1,0.2,0.5\}$. The risk shifts up with $\sigma$ (the constant grows as
    $\sigma^{2.0}$) while the rate never degrades --- every fitted slope is steeper than $-0.8$. Their
    slight steepening with $\sigma$ is the mixture in \eqref{eq:tradeoff_main} shifting toward its
    estimation term, not a $\sigma$-dependent exponent.}
    \label{fig:noise}
\end{figure}

\subsection{Implementation}
\label{app:implementation}

\paragraph{Practical choices in the implementation.}
We fit \eqref{eq:erm} over \eqref{eq:sieve} with cubic B-splines. A few aspects of the fit are settled
on practical grounds rather than read off \eqref{eq:sieve}, and are worth recording together; none of
them weakens what the experiments check.

Several narrow the class rather than widen it, so a bound proved over \eqref{eq:sieve} covers what we
fit. The outer knots sit on the observed range of $T_q(X)$ and are refreshed during fitting, as the
reference implementation does, rather than on the fixed $\mathcal{I}$: a grid fixed on $\mathcal{I}$
wastes most of its resolution, since $\mathcal{I}$ is sized for the worst case, and the observed range
is always a subinterval of it. Multiplicative nodes are parameterized in the log domain, which
constrains $\psi_{qj} > 0$ and so fits the positive-factor subclass (Section~\ref{sec:sim}); the
multiplicative target is built from positive factors and still lies in it. And each target is fitted
with the node-type pattern matching it, one element of the union $\bigcup_\tau \mathcal{F}_n^\tau$;
the over-specified hybrid row of Table~\ref{tab:slopes}, discussed above, is the closest the
experiments come to not being told the pattern.

The rest concern how the boundedness constraints are enforced, and each is measured rather than
assumed. They are imposed in Lagrangian form, by raising a ridge parameter until the coefficients
comply, and the outer bound is the data-dependent $3\max_i |Y_i| + 1$ rather than a fixed $M_g'$;
Remark~\ref{rem:bounded} reports how often each binds and refits every configuration with them
switched off, the refitted slopes moving by at most $0.11$, so no reported convergence slope is an
artifact of the regularization. The derivative constraint $\|g_q'\|_\infty \le L'$ is not imposed
explicitly and does not need to be: on a mesh of width $|\mathcal{I}|/k$ with bounded coefficients it
holds with $L' = O(k)$, the regime the proof of Lemma~\ref{lem:entropy} prices at one further
$\log k_n$ in the entropy --- negligible at the knot counts these fits use, which never exceed $41$.

\paragraph{Optimization.}
The multiplicative fit is the one non-convex piece of the estimator and a minority of starts stall, so
we restart adaptively until the training loss reaches the empirical minimum. The median is used rather
than the mean to summarize replications for exactly this reason: it is the robust summary of the
typical risk and shares the mean's rate exponent when the bulk of runs succeed. Attained training MSE
is recorded for every fit and sits at the noise level $\sigma^2 = 2.5\times 10^{-3}$ throughout the
convergence experiments.

\paragraph{Reproducibility.}
Replication counts are $10$ for the convergence, smoothness and additional-baseline panels, $8$ for
the knot, noise and dimension sweeps, $6$ for the cross-validation panel, and $3$ for the noiseless
fits used above, the constraint measurements of Remark~\ref{rem:bounded} and the $w_0$ sweep above.
Every number quoted anywhere in this paper is either arithmetic or read from one of the stored
per-replication CSVs; nothing is carried over from a scratch calculation.
Every configuration is seeded deterministically from a CRC of the configuration tuple, so seeds do not
depend on the process, the machine, or which subset of experiments is run: re-running one panel with a
different estimator list leaves the other rows bit-identical. Figures are regenerated from the stored
per-replication CSVs rather than by refitting, so a change of rendering never perturbs a reported
number.

\bibliography{references}

\end{document}